\documentclass[journal]{IEEEtran}
\usepackage{amsmath,amsfonts}
\usepackage{algorithmic}
\usepackage{array}
\usepackage[caption=false,font=normalsize,labelfont=sf,textfont=sf]{subfig}
\usepackage{textcomp}
\usepackage{stfloats}
\usepackage{url}
\usepackage{verbatim}
\usepackage{graphicx}
\usepackage{cite}
\usepackage{makecell}
\usepackage{array}%需要该宏包
\usepackage{multirow}
\usepackage{graphbox}
\usepackage{endnotes}
\usepackage{indentfirst}
\usepackage{changepage}
\usepackage{bm}
\usepackage{bbding}
\usepackage{tikz}

\def\BibTeX{{\rm B\kern-.05em{\sc i\kern-.025em b}\kern-.08em
    T\kern-.1667em\lower.7ex\hbox{E}\kern-.125emX}}
\usepackage{balance}
\begin{document}
\newcolumntype{L}[1]{>{\raggedright\arraybackslash}p{#1}}
\newcolumntype{C}[1]{>{\centering\arraybackslash}p{#1}}
\newcolumntype{R}[1]{>{\raggedleft\arraybackslash}p{#1}}

\title{Cross-Spatial Pixel Integration and Cross-Stage Feature Fusion Based Transformer Network for Remote Sensing Image Super-Resolution}
\author{Yuting Lu, Lingtong Min,
        Binglu Wang$\dagger$,~\IEEEmembership{Member,~IEEE,}
        Le Zheng,~\IEEEmembership{Senior Member,~IEEE,}\\
	  Xiaoxu Wang,~\IEEEmembership{Member,~IEEE,}     
		Yongqiang Zhao,~\IEEEmembership{Member,~IEEE,}
            and Teng Long,~\IEEEmembership{Fellow,~IEEE}\\
		% <-this % stops a space
        \thanks{Yuting Lu, Xiaoxu Wang and Yongqiang Zhao are with School of Automation, Northwestern Polytechnical University, Xi’an 710072, China (e-mail:lyt1996@mail.nwpu.edu.cn, woyaofly1982@nwpu.edu.cn, zhaoyq@nwpu.edu.cn). Lingtong Min is with School of Electronics and Information, Northwestern Polytechnical University, Xi’an 710072, China (e-mail:minlingtong@nwpu.edu.cn).}
		\thanks{Binglu Wang, Le Zheng and Teng Long are with the Radar Research Laboratory, School of Information and Electronics, Beijing Institute of Technology,Beijing 100081, China  (e-mail: wbl921129@gmail.com, le.zheng.cn@gmail.com, longteng@bit.edu.cn).}
      
   \thanks{$\dagger$Corresponding author: Binglu Wang.}
   % <-this % stops a space
		% <-this % stops a space        
		%\thanks{}% <-this % stops a space
		% <-this % stops a space
		\thanks{This work is supported by the Postdoctoral Science Foundation of China under Grant 2022M710393, the Fourth Special Grant of China Postdoctoral Science Foundation (in front of the station) 2022TQ0035 and the Shaanxi Science Fund for Distinguished Young Scholars 2022JC-49.}
  }

\markboth{Journal of \LaTeX\ Class Files,~Vol.~18, No.~9, September~2020}%
{How to Use the IEEEtran \LaTeX \ Templates}

\maketitle

\begin{abstract}
Remote sensing image super-resolution (RSISR) plays a vital role in enhancing spatial detials and improving the quality of satellite imagery. Recently, Transformer-based models have shown competitive performance in RSISR. To mitigate the quadratic computational complexity resulting from global self-attention, various methods constrain attention to a local window, enhancing its efficiency. Consequently, the receptive fields in a single attention layer are inadequate, leading to insufficient context modeling. Furthermore, while most transform-based approaches reuse shallow features through skip connections, relying solely on these connections treats shallow and deep features equally, impeding the model's ability to characterize them. To address these issues, we propose a novel transformer architecture called Cross-Spatial Pixel Integration and Cross-Stage Feature Fusion Based Transformer Network (SPIFFNet) for RSISR. Our proposed model effectively enhances global cognition and understanding of the entire image, facilitating efficient integration of features cross-stages. The model incorporates cross-spatial pixel integration attention (CSPIA) to introduce contextual information into a local window, while cross-stage feature fusion attention (CSFFA) adaptively fuses features from the previous stage to improve feature expression in line with the requirements of the current stage. We conducted comprehensive experiments on multiple benchmark datasets, demonstrating the superior performance of our proposed SPIFFNet in terms of both quantitative metrics and visual quality when compared to state-of-the-art methods.
\end{abstract}


\section{Introduction}
\IEEEPARstart{R}{emote} sensing imaging technology is of paramount importance in numerous fields, including environmental monitoring \cite{yin2021cascaded, yang2015compressive, xue2021multilayer}, disaster management \cite{min2023d, wang2022local}, urban planning \cite{zhang2022coupling, yang2021variational}, and object detection \cite{liu2023distilling, xue2019nonlocal, yang2023unsupervised}. Therefore, the acquisition of high-resolution remote sensing images is imperative for the effective implementation and analysis of remote sensing image applications. However, challenges arise due to factors such as sensor noise, optical distortion and environmental interference, which can significantly degrade the image quality. Image super-resolution (SR) is a typical computer vision task that involves reconstructing high-resolution (HR) images from low-resolution (LR) images. The primary objective of SR is to mitigate the detrimental impact of acquisition equipment and environmental factors on remote sensing imaging outcomes, thereby enhancing the resolution of remote sensing images. As an alternative to developing physical imaging technologies, SR has gained significant attention in recent years for its ability to effectively generate high-resolution remote sensing images \cite{wang2022review}.

Traditional RSISR methods often rely on interpolation-based techniques, such as bicubic interpolation \cite{keys1981cubic} or Lanczos interpolation. While these methods are simple, they may yield limited performance due to their inability to capture high-frequency details and structural information in the generated images. Recent advancements in deep learning \cite{lecun2015deep} have led to the emergence of convolutional neural networks (CNNs) as powerful tools for various image processing tasks, including RSISR. CNN-based methods \cite{jia2022multiattention, lei2019coupled, dong2020remote1, wang2021contextual, jin2023learning, liu2022model, li2023progressive} have demonstrated promising results in learning complex representations from large datasets. However, despite the success of CNNs, they still possess certain limitations when employed for RSISR. CNNs typically operate locally with fixed receptive fields, which may hinder their ability to effectively capture long-range dependencies. As a result, they may have limited modeling capacity for remote sensing scenes with large spatial extent \cite{he2023deep}.\\
\begin{figure*}[htp]
    \centering
    \includegraphics[width=1\linewidth]{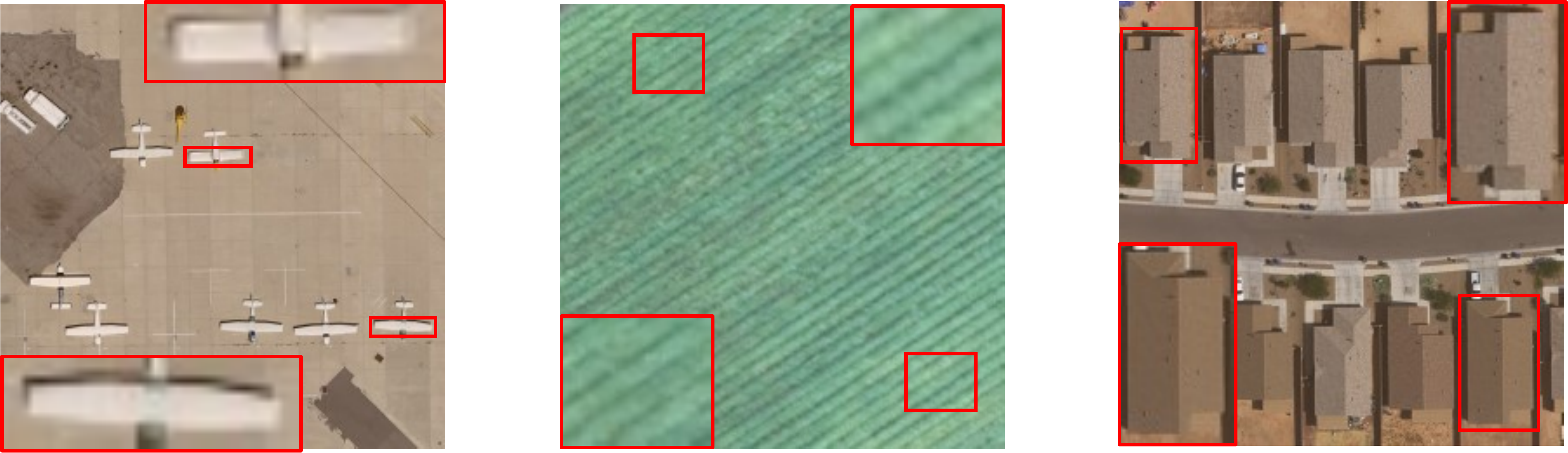}
    \caption{Examples of similar windows in remote sensing images. These windows are similar in texture, shape and semantic features but far apart in space.}
    \label{fig:1}
\end{figure*}
\begin{figure}[t]
    \centering
    \includegraphics[width=1\linewidth]{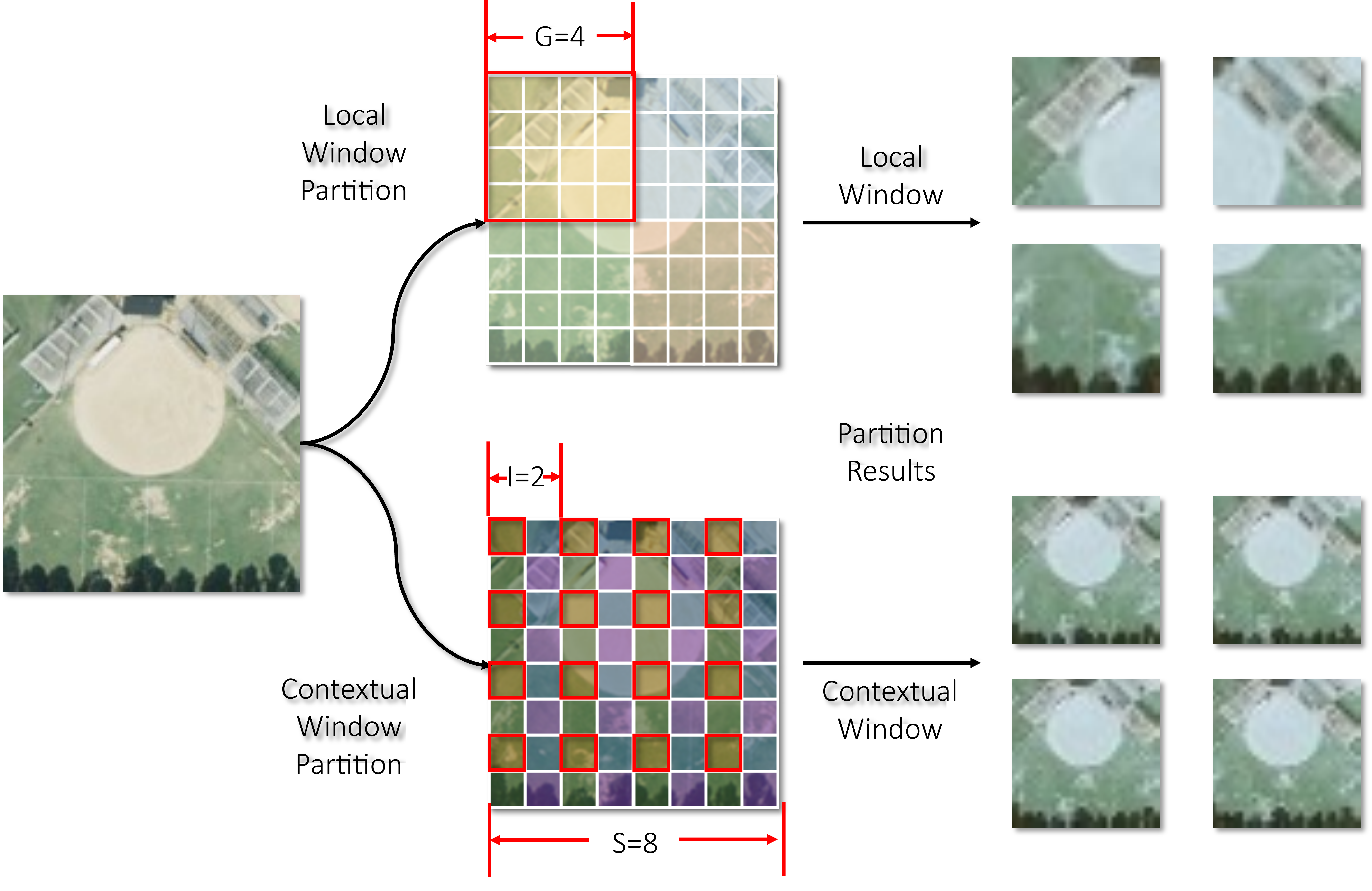}
    \caption{Diagrams of difference between local window partition and contextual window partition. The pixels in the red box belong to the same window.}
    \label{fig:difference}
\end{figure}
\indent Transformer-based architectures, initially developed for natural language processing tasks, have emerged as a suitable solution for addressing the limitations of CNNs and have demonstrated impressive performance in diverse computer vision tasks, including image classification \cite{dosovitskiy2020image, wang2021pyramid, wu2021cvt} and object detection \cite{carion2020end, beal2020toward, sun2021rethinking}. The pioneering vision transformer model \cite{dosovitskiy2020image} employs a redundant attention mechanism, resulting in quadratic computation complexity relative to the image size. This high computational complexity poses challenges for its application in high-resolution predictions for RSISR tasks. To mitigate this issue, recent proposals have explored the use of self-attention within small spatial regions \cite{chen2021pre, liang2021swinir, wang2022uformer}. However, since remote sensing images usually cover a large range of areas, ground objects and landforms with similar features are far apart in space. As shown in Fig. \ref{fig:1}, the texture, shape and semantic features of the areas in the boxes are similar to each other but far apart in space. As a result, these methods that partition an image into fixed-size windows and employ self-attention within those windows to model pixel dependencies fail to capture interactions between distant but similar pixels, which is crucial for attaining optimal performance \cite{lam}. Furthermore, the majority of current transformer-based RSISR methods primarily rely on skip connections to transmit shallow features to deep features. However, treating these reused shallow features equally impedes the representational capability of transformers, despite the proven effectiveness of skip connections in RSISR \cite{rcan, lan2020cascading, lan2020madnet}.

\begin{figure*}[htp]
    \centering
    \includegraphics[width=1\linewidth]{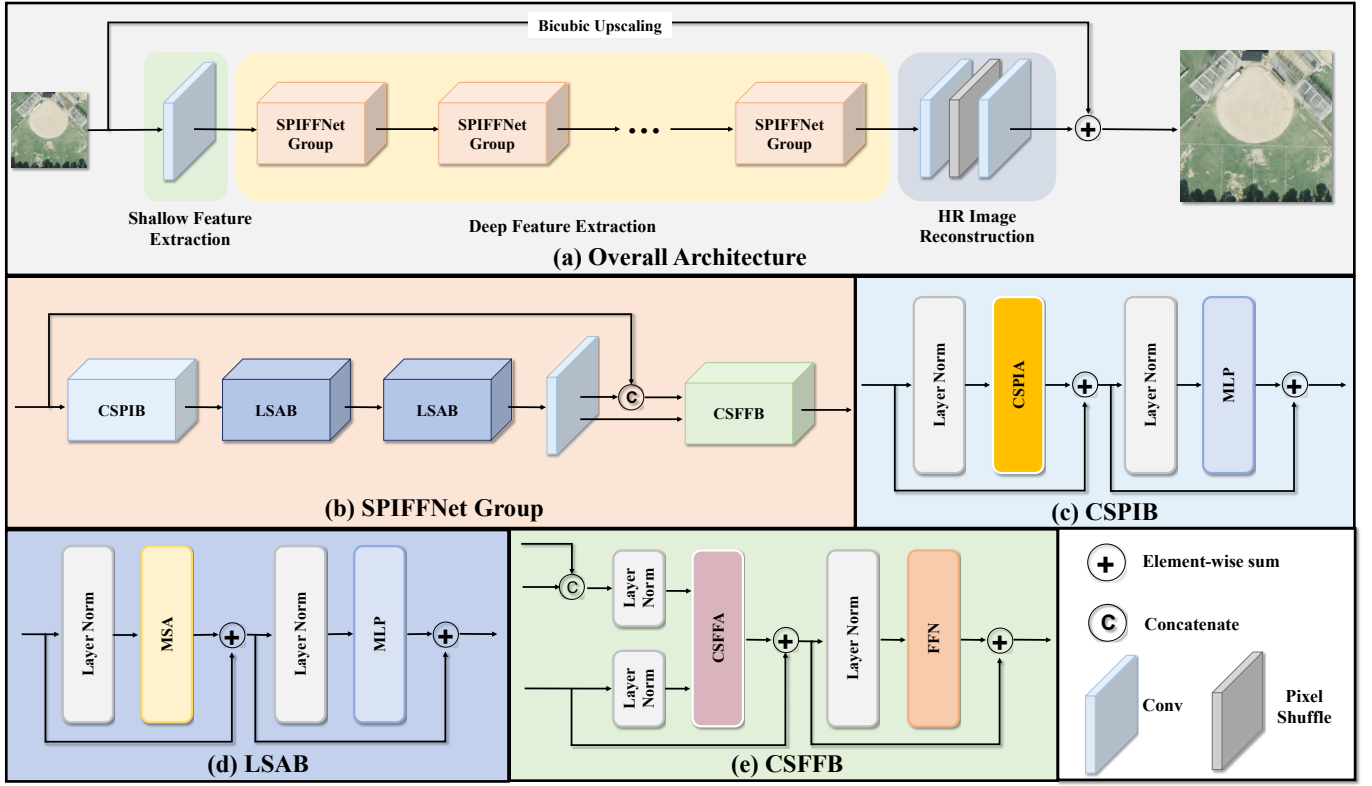}
    \caption{(a) Architecture of SPIFFNet for high-resolution image restoration. (b) The main components of the model: Cross-Spatial Pixel Integration Block (CSPIB), Local Spatial Attention Block (LSAB), Local $3 \times 3$ Convolution and Cross-Stage Feature Fusion Block (CSFFB). (c) Cross-Spatial Pixel Integration Block (CSPIB) that implements the injection of context information into a local window. (d) Local Spatial Attention Block (LSAB) model local-range dependencies in the fine area of a local window. (e) Cross-Stage Feature Fusion Block (CSFFB) performs channel attention across stages for feature fusion.}
    \label{fig:Overview}
\end{figure*}

To address these limitations, we propose two components: cross-space pixel Fusion attention (CSPIA) and cross-stage feature fusion attention (CSFFA). CSPIA  allows the local window to perceive the contextual window (refer to Fig. \ref{fig:difference}) by maximizing the similarity between image pairs. This, in turn, effectively enlarges the receptive field, as depicted in Fig. \ref{fig:ganshouye}, enabling the utilization of valuable context information from the image. In parallel, CSFFA enhances feature expression by adaptively integrating features cross-stages. CSPIA consists of three main steps: Space Division (SD), Local-Context Matching (LCM) and Cross Attention (CA). The SD is responsible for obtaining local windows and contextual windows through different spatial partitioning strategies. Then, LCM is used to obtain the most matched contextual window for each pair of local window. Finally, the most similar contextual window is selected to conduct CA with corresponding local window. In this way, context information can be integrated into current local window efficiently, as shown in Fig \ref{fig:CSPIA}. Building upon CSPIA, we construct a cross-spatial pixel integration block (CSPIB). Following the fusion of pixels from contextual windows, we apply the standard multihead self-attention (MSA) to capture local-range dependencies within the refined area of the local window and a local $3 \times 3$ convolution further handles local details. Subsequently, CSFFA calculates cross-covariance of feature channels across stages to generate cross-stage attention map based on both shallow features and deep features (after projection of key and query). CSFFA enables the model to adaptively adjust the channel-wise feature maps at cross-stage of the network to enhance the informative multiscale feature representation ability. Furthermore, to enhance the flow of complementary features and allow subsequent network layers to focus on finer image details, we integrate a feed forward network (FFN) \cite{zamir2022restormer} into our model. Utilizing CSFFA and FFN, we construct a cross-stage feature fusion block (CSFFB). By combining CSPIBs and CSFFBs, we develop a transformer network, named SPIFFNet, which incorporates cross-spatial pixel integration and cross-stage feature fusion, specifically designed for RSISR. This architecture is depicted in Fig. \ref{fig:Overview}. Furthermore, our experimental findings unequivocally demonstrate the superiority of the proposed SPIFFNet model over state-of-the-art methods.

The article presents three key contributions that can be summarized as follows:
\begin{figure}[t]
    \centering
    \includegraphics[width=1\linewidth]{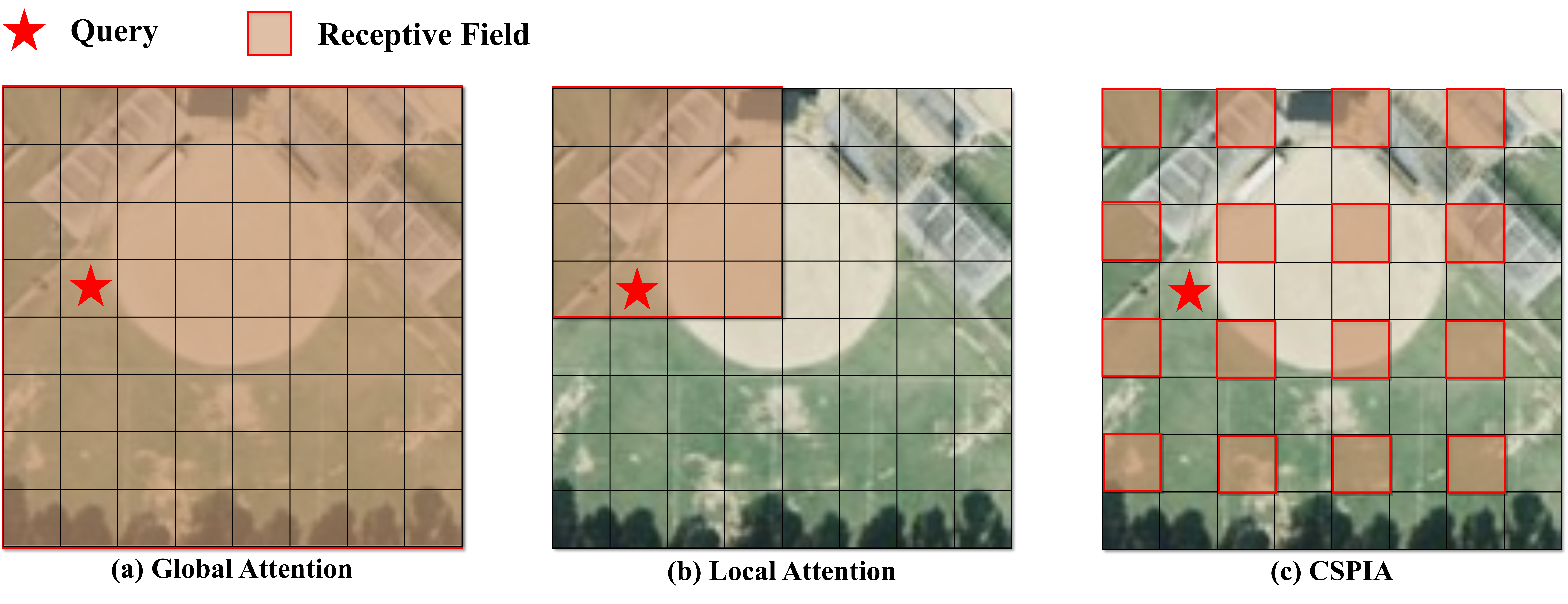}
    \caption{Comparison of CSPIA with other Transformer models. The red star denote the query, and masks with solid line boundaries denote the regions to which the queries attend. (a) Global Attention \cite{transenet} adopts full attention for all queries. (b) Local Attention \cite{liang2021swinir} uses partitioned window attention. (c) CSPIA learns the context information region that is most similar to current query.}
    \label{fig:ganshouye}
\end{figure}

1) We propose cross-spatial pixel integration attention (CSPIA) to introduce contextual information into the local window. The contextual information of the image enhances the global cognition and understanding of the entire image. By incorporating the context information within the local window, the model gains a better understanding of the relationship between ground features and the surrounding environment, thereby enhancing the consistency and accuracy of the RSISR results.

2) We propose cross-stage feature fusion attention (CSFFA), which facilitates effective feature representation by modeling the interdependencies among different channels across stages. By dynamically assigning weights to different channels, this mechanism enhances the model's capacity to capture essential image features while suppressing irrelevant ones, thereby producing higher-quality super-resolved images.

3) Based on CSPIA and CSFFA, we propose SPIFFNet, a cross-spatial pixel integration and cross-stage feature fusion based transformer network for remote sensing image super-resolution. SPIFFNet effectively captures contextual information to enhance the global perception ability of the local window and adaptively fuses information from previous stages to enhance feature representation. Experimental results on benchmark datasets validate that SPIFFNet achieves state-of-the-art performance in terms of objective metrics as well as visual quality.

The rest of this article is organized as follows. Section \uppercase\expandafter{\romannumeral 2} presents the related works on SR. Section \uppercase\expandafter{\romannumeral 3} introduces the proposed SPIFFNet model, including the CSPIA and the CSFFA. Section \uppercase\expandafter{\romannumeral 4} presents the experimental results and analysis. Finally, Section \uppercase\expandafter{\romannumeral 5} concludes this paper.
\section{RELATED WORK}
\subsection{Deep Learning-based Methods for SR}
Deep learning-based super-resolution (SR) methods predominantly rely on standard convolutional neural networks (CNNs) owing to their robust nonlinear representation capabilities. Typically, these methods approach super-resolution as an image-to-image regression task, with the objective of learning the direct mapping from LR to HR images. SRCNN \cite{srcnn} first uses three convolution layers to map the low-resolution images to high-resolution images. Building upon SRCNN, Kim \emph{et al}. extended the network depth in their work called DRCN \cite{drcn}, resulting in considerable performance improvements over SRCNN \cite{srcnn}. FSRCNN \cite{fsrccc} achieves high computational efficiency without compromising restoration quality through a redesigned architecture of SRCNN. VDSR \cite{VDSR} addressed the challenge of handling multi-scale images within a unified framework by incorporating residual learning, gradient cropping, and an increased number of network layers. EDSR \cite{edsr} achieves superior performance by streamlining the model architecture, eliminating redundant modules from the conventional ResNet framework. For remote sensing images, LGCNet \cite{LGCnet} stands as the pioneering CNN-based model for super-resolution, introducing the concept of local and global contrast features to enhance the preservation of details and clarity in the reconstructed images. Haut \emph{et al}. \cite{dcm} coordinates several different improvements in network design to achieve the most advanced performance on the RSISR task. A novel single-path feature reuse approach and a second-order learning mechanism are proposed by Dong \emph{et al}. \cite{dong2020remote}, which aim to effectively utilize both small and large difference features. Although these methods have achieved impressive results, the limited receptive field of CNNS cannot capture the long-range dependencies between pixels, thereby limiting their performance.
\subsection{Transformer-based Methods}
The Transformer network, initially proposed in 2017 for machine translation tasks \cite{vaswani2017attention}, has gained popularity in computer vision due to its remarkable performance in image processing \cite{dosovitskiy2020image, parmar2018image}. Since its inception, numerous visual models based on the Transformer have been proposed \cite{chen2021pre, wang2022uformer, zamir2022restormer, lu2022transformer, transenet}. As an example, Chen \emph{et al}. introduced the Image Processing Transformer (IPT) as a novel pre-trained model for low-level computer vision tasks. To fully leverage the potential of the transformer, a substantial amount of corrupted image pairs is generated using the ImageNet dataset. The IPT model adapts to diverse image processing tasks through multi-head and multi-tail training, along with the incorporation of contrastive learning techniques. Another well-known image restoration method called Uformer \cite{wang2022uformer} utilizes the Locally Enhanced Window Transformer block, which reduces computational requirements by utilizing a non-overlapping window-based self-attention mechanism. Moreover, three skip connection schemes are explored to facilitate efficient information transfer from the encoder to the decoder. Furthermore, the Restoration Transformer (Restormer) network \cite{zamir2022restormer} proposes an effective Transformer model that captures remote pixel interactions and is suitable for large image, which is achieved through key design choices in the building blocks, including multi-head attention and feedforward networking. Efficient Super-Resolution Transformer (ESRT) \cite{lu2022transformer} presents a hybrid model that combines a lightweight CNN backbone (LCB) with a lightweight transformer backbone (LTB) can dynamically adjusts the feature map size, achieving competitive results with low computational cost. SwinIR \cite{liang2021swinir} introduced a robust baseline model for image restoration that utilizes the Swin Transformer architecture. In the context of RSISR, TransENet \cite{transenet} proposes a multilevel enhancement architecture based on the Transformer framework, which can be integrated with the conventional super-resolution (SR) framework to effectively merge multi-scale high- and low-dimensional features. 
\section{METHODOLOGY}
In this section, we introduce the proposed SPIFFNet for RSISR. The overall framework of SPIFFNet is presented in Section III-A and the SPIFFNet gruop that integrates CSPIB and CSFFB is carefully discussed in Section III-B. Furthermore, in Section III-C, we will provide a concise overview of the implementation details.
\subsection{Overview of SPIFFNet}
In this section, we introduce the framework of our method, as shown in Fig. \ref{fig:Overview}. Given an input image ${I_{LR}\in {\mathbb R}^{H \times W \times 3}}$, where $H$ and $W$ are the image height and width. Then, the input ${I_{LR}}$ undergoes a transformation into the feature space through a $3 \times 3$ convolutional layer
\begin{equation}\label{eqn-1} 
{F_0={\rm Conv}(I_{LR})} 
\end{equation}
where the Conv denotes $3 \times 3$ convolution and the ${F_0} \in {\mathbb R}^{H \times W \times C}$ represents the shallow features.
\begin{figure*}[htp]
    \centering
    \includegraphics[width=1\linewidth]{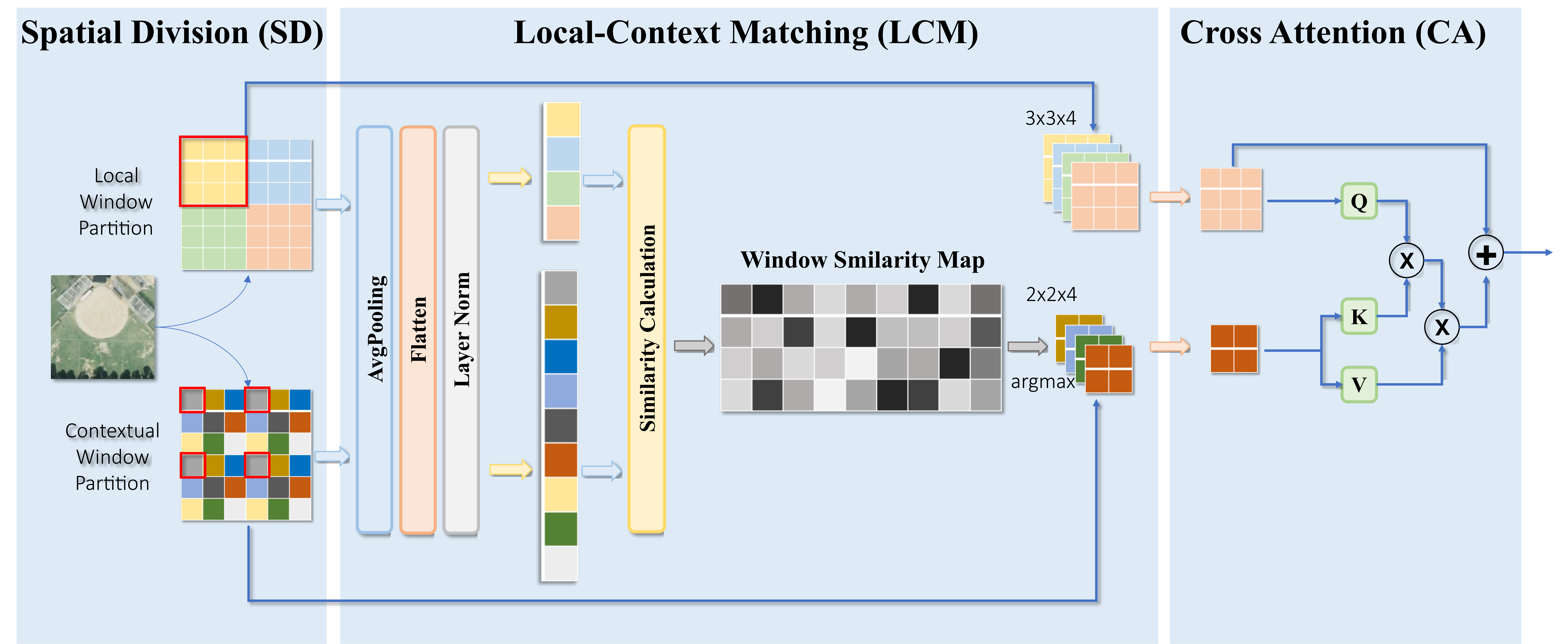}
    \caption{Illustration of cross-spatial pixel integration attention (CSPIA). \textcircled{×} and \textcircled{+} are matrix multiplication and element-wise addition operations, respectively.}
    \label{fig:CSPIA}
\end{figure*}

Then, several SPIFFNet groups, each of which involves CSPIB, LSAB, local $3 \times 3$ convolution and CSFFB are set up after the convolutional layer for deep feature extraction. We extract deep feature ${F_{DF}\in {\mathbb R}^{H \times W \times C}}$ from ${F_0}$ as
\begin{equation}\label{eqn-2}
{F_{DF}={H_{DF}(F_0)}}
\end{equation}
where ${H_{DF}}$ represents the deep feature extraction module which contains $K$ SPIFFNet groups. Specifically, the intermediate features $F_1, F_2, \ldots, F_K$ and the final deep feature $F_{DF}$ are extracted sequentially
\begin{equation}\label{eqn-3}
{F_{i}={H_{i}(F_{i-1})}, \qquad i=1,2,...,K} 
\end{equation}
where $H_{i}$ denotes the $i$-th SPIFFNet group.

Finally, the deepest features $F_{DF}$ are reconstructed using a $3 \times 3$ convolutional layer and pixel-shuffle upsampling operations \cite{pixel-shuffle} to generate SR image $I_{r}$. In addition, a bilinear interpolation of the LR image $I_{b}$ is incorporated in the summation process to aid in the recovery process for the super-resolution output $I_{r}$
\begin{equation}\label{eqn-4}
{I_{SR}=I_{r}+I_{b}}
\end{equation}

We train the proposed model using the L1 loss function. The loss function is obtained by comparing the LR images $I_{LR}$ with their corresponding HR reference images $I_{HR}$ 
\begin{equation}\label{eqn-5}
L_{(\theta)}=\frac{{\rm{1}}}{N}\sum\limits_{i = 1}^N {{{\left\| {I_{HR}^{(i)} - I_{SR}^{(i)}} \right\|}_1}} 
\end{equation}
where ${\theta}$ represents the parameters of the SPIFFNet, and $N$ denotes the number of training samples.
\subsection{SPIFFNet Group}
In this section, we introduce the SPIFFNet group, a crucial component of our SPIFFNet model. Each SPIFFNet group consists of four components: cross-spatial pixel integration attention block (CSPIB), Local Spatial Attention Block (LSAB), local $3 \times 3$ convolution and cross-stage feature fusion block (CSFFB). The Cross-Spatial Pixel Integration Block (CSPIB) expands the model's receptive field, allowing it to capture long-range dependencies and contextual information from the input feature maps. This expansion is facilitated by the utilization of the Cross-Spatial Pixel Integration Attention (CSPIA). The LSAB is responsible for capturing the correlations of local spatial information. Local $3 \times 3$ convolution deals with local details in a fine-grained manner. CSFFB is designed to adaptive integration of information from the previous stage according to the needs of the characteristics of the current stage rather than treating them equally. This combination of local and global information, along with the cross stage adaptive information fusion, enables the model to capture complex spatial dependencies and contextual information effectively.

\emph{1) Cross-Spatial Pixel Integration Block (CSPIB):} Previous methods often overlooked the interaction between local and contextual features. The CSPIB is designed to expand the local spatial window to capture more context information, as shown in Fig. \ref{fig:Overview}(c). CSPIB contains two sequential modules, the cross-spatial pixel integration attention (CSPIA) is designed to capture contextual information for local windows and the MLP module for feature projection.

\textbf{Cross-Spatial Pixel Integration Attention (CSPIA):} In this section, our objective is to expand the receptive field of local windows, enabling them to capture context information from the input feature maps. Previous studies have demonstrated that SR networks with a wider effective receptive field achieve superior performance \cite{lam}. The challenge lies in enabling the network to model global connectivity while preserving computational efficiency. Due to the fixed partitioning of windows at the layer level, there are no direct connections between windows. One straightforward approach is to exhaustively combine the information from every window pair. However, this approach is unnecessary and inefficient since many windows are irrelevant and uninformative. Additionally, redundant interactions may introduce noise that impairs the model's performance. Based on these observations, we introduce an innovative technique called cross-spatial pixel integration attention (CSPIA), where each local window adaptively integrates pixels with the most correlated global window. Specifically, as shown in Fig. \ref{fig:CSPIA}, the CSPIA consists of three steps: Spatial Division (SD), Local-Context Matching (LCM) and Cross Attention (CA). Through SD, we split the feature map into two parts: local windows and global windows. In the case of local windows, adjacent embeddings of size $G \times G$ are grouped together. An example is illustrated in Fig. \ref{fig:difference} with $G = 4$. In the case of global windows, where the input size is $S \times S$, the feature map is sampled at a fixed interval $I$. Fig. \ref{fig:difference} demonstrates an example with $I = 2$, where embeddings with a red border belong to a window. The height or width of the group for global windows is calculated as $G = S/I$. All windows can be processed in parallel, after which the outputs are pasted to their original location in window aggregation module. It is worthwhile to note that such cropping strategy is adaptive to arbitrary input size, which means no padding pixels are needed. Then, local windows and contextual windows are spatially pooled into one-dimensional tokens. These tokens encode the characteristic of the windows, which are later used for similarity calculation and window matching. This process can be expressed as 
\begin{equation}\label{eqn-6}
{{\bar X}_i} = \mathop {\arg \max }\limits_{{X_j}} L{({X_i})^T}L({X_j}),{\rm{   }}j \ne i
\end{equation}
where ${\bar X}_i$ is the best-matching global window with current local window $X_i$, and $L( \cdot )$ is the average pooling function along spatial dimension followed by flatten operation and layer normalization. Since the \emph{argmax} opration is non-differentiable, we replace it with Gumbel-Softmax opration \cite{jang2016categorical} during training so as to make it possible to train end-to-end. After that, pixel information of ${\bar X}_i$ are fused into $X_i$ via Cross-Attention (CA) 
\begin{equation}\label{eqn-7}
{X_i} = CA({X_i},{{\bar X}_i})
\end{equation}
As illustrated in Fig. \ref{fig:CSPIA}, CA works in a similar way to the standard self-attention \cite{dosovitskiy2020image}, but the key and value are calculated using ${\bar X}_i$. As a result, CSPIA can enable contextual pixel integration while introducing little computational overhead.
\begin{figure*}[htp]
    \centering 
    \includegraphics[width=1\linewidth]{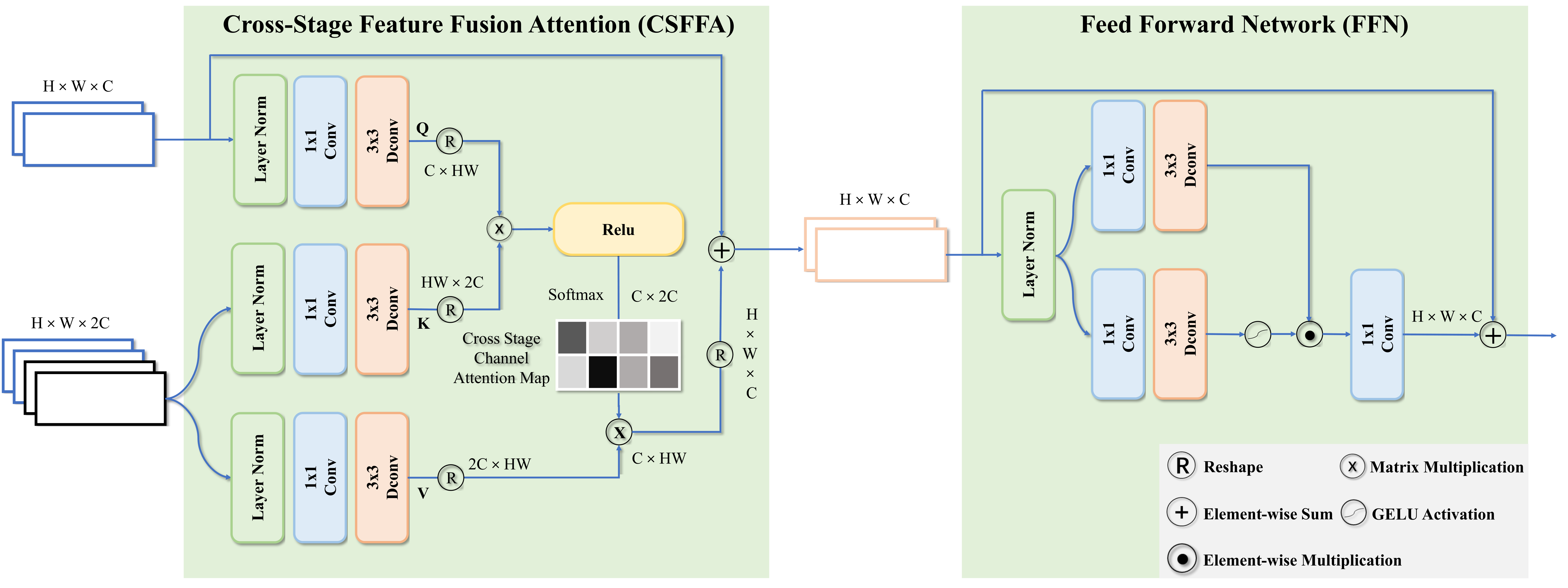}
    \caption{Illustration of cross-stage feature fusion block (CSFFB). It consists of two parts: cross-stage feature fusion attention (CSFFA) and FFN.}
    \label{fig:CSFFA}
\end{figure*}

\emph{2) Local Spatial Attention Block (LSAB):} As shown in Fig. \ref{fig:Overview}(d), LSAB adopts the standard multihead self-attention (MSA) paradigm \cite{dosovitskiy2020image}, with two modifications. Firstly, LSAB operates at the window level instead of the image level. Secondly, positional embedding is omitted due to the introduction of the convolutional layer, which implicitly learns positional relationships and enhances the network's efficiency and conciseness. LSAB is designed to model local-range dependencies within a window, facilitating the comprehensive utilization of contextual information.

Specifically, for feature $X \in {{\mathbb R}^{{P^2} \times C}}$, the corresponding \emph{query}, \emph{key} and \emph{value} matrices $Q \in {{\mathbb R}^{{P^2} \times d}}$, $K \in {{\mathbb R}^{{P^2} \times d}}$, $V \in {{\mathbb R}^{{P^2} \times C}}$ are computed as 
\begin{equation}\label{eqn-8}
Q = X{W_Q},{\quad}K = X{W_K},{\quad}V = X{W_V}
\end{equation}
where the weight matrices $W_Q$, $W_K$ and $W_V$ are shared across windows, $P$ is the window size. By comparing the similarity between $Q$ and $K$, we obtain a attention map of size ${\mathbb R}^{P^2 \times P^2}$ and multiply it with $V$. Overall, the calculation of Multi-head Self-Attention (MSA) can be expressed as 
\begin{equation}\label{eqn-9}
MSA(X) = softmax (Q{K^T}/\sqrt d )V
\end{equation}
Here $\sqrt d $ is used to control the magnitude of $Q{K^T}$ before applying the softmax function.

Similar to the conventional transformer layer \cite{dosovitskiy2020image}, the MLP is employed after MSA module to further transform features. MLP contains two fully-connected layers, and one GELU nonlinearity is applied after the first linear layer.

\emph{3) Local $3 \times 3$ Convolution:} By adding a local $3 \times 3$ convolutional layer after feature extraction, the Transformer-based network is infused with the inherent inductive bias of convolution operations. This enhances the foundation for aggregating shallow and deep features in subsequent stages.

\emph{4) Cross-Stage Feature Fusion Block (CSFFB):} Skip Connections are commonly used to propagate shallow features to deeper layers. However, the long-term information from the shallow stages tends to be attenuated. Although the shallow features can be reused through skip connections, they are treated indiscriminately with the deep features across different stages, thereby impeding the representational capacity of CNNs. To address this concern, we introduce CSFFB, illustrated in Fig. \ref{fig:Overview}(e). CSFFB comprises two consecutive components: the Cross-Stage Feature Fusion Attention (CSFFA) for adaptive feature fusion across stages, and the FFN for feature transformation.
\begin{table*}[]
\centering
\caption{MEAN PSNR (DB) AND SSIM OVER THE UCMERCED TEST DATASET AND THE AID TEST DATASET}
\label{tab:all}
% \resizebox{1\linewidth}{!}{
\setlength{\tabcolsep}{3.6mm}{%
\begin{tabular}{cllccccccc}
\hline
\textbf{Dataset} & \multicolumn{1}{c}{\textbf{Scale}} & \multicolumn{1}{c}{\textbf{Metric}} & \textbf{SRCNN} & \textbf{LGCNet} & \textbf{VDSR} & \textbf{DCM} & \textbf{HSENet} & \textbf{TransENet} & \textbf{\begin{tabular}[c]{@{}c@{}}SPIFFNet\\ Ours\end{tabular}} \\ \hline
\multicolumn{1}{c|}{\multirow{6}{*}{\textbf{UCMerced}}} & \multirow{2}{*}{\textbf{×2}} & \textbf{PSNR} & 32.84 & 33.48 & 33.87 & 33.65 & 34.22 & 34.03 & \textbf{34.68} \\
\multicolumn{1}{c|}{} &  & \textbf{SSIM} & 0.9152 & 0.9235 & 0.9280 & 0.9274 & 0.9327 & 0.9301 & \textbf{0.9354} \\ \cline{2-10} 
\multicolumn{1}{c|}{} & \multirow{2}{*}{\textbf{×3}} & \textbf{PSNR} & 28.66 & 29.28 & 29.76 & 29.52 & 30.00 & 29.92 & \textbf{30.43} \\
\multicolumn{1}{c|}{} &  & \textbf{SSIM} & 0.8038 & 0.8238 & 0.8354 & 0.8394 & 0.8420 & 0.8408 & \textbf{0.8510} \\ \cline{2-10} 
\multicolumn{1}{c|}{} & \multirow{2}{*}{\textbf{×4}} & \textbf{PSNR} & 26.78 & 27.02 & 27.54 & 27.22 & 27.73 & 27.77 & \textbf{28.09} \\
\multicolumn{1}{c|}{} &  & \textbf{SSIM} & 0.7219 & 0.7333 & 0.7522 & 0.7528 & 0.7623 & 0.7630 & \textbf{0.7733} \\ \hline
\multicolumn{1}{c|}{\multirow{6}{*}{\textbf{AID}}} & \multirow{2}{*}{\textbf{×2}} & \textbf{PSNR} & 34.49 & 34.80 & 35.05 & 35.21 & 35.30 & 35.28 & \textbf{35.66} \\
\multicolumn{1}{c|}{} &  & \textbf{SSIM} & 0.9286 & 0.9320 & 0.9346 & 0.9366 & 0.9377 & 0.9374 & \textbf{0.9397} \\ \cline{2-10} 
\multicolumn{1}{c|}{} & \multirow{2}{*}{\textbf{×3}} & \textbf{PSNR} & 30.55 & 30.73 & 31.15 & 31.31 & 31.39 & 31.45 & \textbf{31.70} \\
\multicolumn{1}{c|}{} &  & \textbf{SSIM} & 0.8372 & 0.8417 & 0.8522 & 0.8561 & 0.8581 & 0.8595 & \textbf{0.8631} \\ \cline{2-10} 
\multicolumn{1}{c|}{} & \multirow{2}{*}{\textbf{×4}} & \textbf{PSNR} & 28.40 & 28.61 & 28.99 & 29.17 & 29.34 & 29.38 & \textbf{29.54} \\
\multicolumn{1}{c|}{} &  & \textbf{SSIM} & 0.7561 & 0.7626 & 0.7753 & 0.7824 & 0.7881 & 0.7909 & \textbf{0.7938} \\ \hline
\end{tabular}%
}
\end{table*}

\textbf{Cross-Stage Feature Fusion Attention (CSFFA):} Figure \ref{fig:CSFFA} demonstrates the operation of the CSFFA, which calculates attention scores at the channel level using feature maps cross-stages. These scores are then applied to the feature maps, enabling the weighting of each channel's contribution and the integration of features from previous stages with the current input. This process promotes the fusion of channel information across stages, facilitating the model to capture both low-level details and high-level contextual information. As a result, the approach enables more precise and effective image super-resolution.

Specifically, the features of the current stage, denoted as $X_{cur} \in \mathbb{R}^{H \times W \times C}$, and the previous stage, denoted as $X_{pre} \in \mathbb{R}^{H \times W \times C}$, are concatenated along the feature dimension to obtain $Y \in \mathbb{R}^{H \times W \times 2C}$. Our CSFFA then gets \emph{query}, \emph{key}, and \emph{value} projections: $Q = W_d^{Q}W_p^{Q}X_{cur}$, $K = W_d^{K}W_p^{K}Y$, and $V = W_d^{V}W_p^{V}Y$. Where $W_p^{( \cdot )}$ denotes the $1 \times 1$ point-wise convolution, and $W_d^{( \cdot )}$ represents the $3 \times 3$ depth-wise convolution. Subsequently, we reshape the $Q$ and $K$ to facilitate their dot-product interaction, resulting in a cross-stage channel attention map $A$ with dimensions $\mathbb{R}^{C \times 2C}$. The CSFFA process can be defined as
\begin{equation}\label{eqn-10}
\begin{aligned}
X_{out} = {W_p}Attn(Q,K,V) + X_{cur} \\
Attn(Q,K,V) = V \cdot Softmax (Relu(Q \cdot K/\alpha ))
\end{aligned}
\end{equation}
where $X_{cur}$ and ${X_{out}}$ are the input and output feature maps, respectively; $Q \in {\mathbb R}^{C \times HW}$; $K \in {\mathbb R}^{HW \times 2C}$; and $V \in {\mathbb R}^{2C \times HW}$ matrices are obtained by the $X_{cur} \in {\mathbb R}^{H \times W \times C}$ and ${Y\in {\mathbb R}^{H \times W \times 2C}}$, respectively. To enhance feature control and promote the development of sophisticated image attributes, we introduce a ReLU non-linearity function before the softmax normalization. The ReLU non-linearity function applies sparse constraints to the cross-stage attention map promotes the model's focus on the most informative regions and mitigates the influence of noisy or irrelevant features.

To transform features, we use two parallel $1 \times 1$ convolutions and $3 \times 3$ convolutions to the feature map. Subsequently, a SimpleGate activation function \cite{chen2022simple} multiplies one of the branches to regulate the flow of complementary features and facilitate feature transformation. To be specific, when provided with an input tensor $X \in {\mathbb R}^{H \times W \times C}$, the FFN can be expressed as
\begin{equation}\label{eqn-11}
\begin{aligned}
\hat X &= W_p^0Gating(X) + X\\
Gating(X) &= \phi (W_d^1W_p^1(LN(X))) \odot W_d^2W_p^2(LN(X))
\end{aligned}
\end{equation}
where the symbol $ \odot $ denotes element-wise multiplication, $\phi$ refers to the GELU non-linearity function, and LN denotes layer normalization \cite{ba2016layer}.
\subsection{Implementation Details}
\begin{table}[]
\centering
\caption{ABLATION EXPERIMENTS ON UCMERCED-X4}
\label{tab:Ablation Studies}
% \resizebox{1\linewidth}{!}{
\setlength{\tabcolsep}{5mm}{%
\begin{tabular}{cc|llllll}
\hline
\multicolumn{1}{l}{\textbf{CSPIA}} & \textbf{CSFFA} & \multicolumn{2}{c}{\textbf{Param}} & \multicolumn{2}{c}{\textbf{PSNR}} & \multicolumn{2}{c}{\textbf{SSIM}} \\ \hline
\textbf{\XSolidBrush} & \textbf{\XSolidBrush} & \multicolumn{2}{c}{1.459M} & \multicolumn{2}{c}{28.009} & \multicolumn{2}{l}{0.7711} \\ \hline
\textbf{\XSolidBrush} & \textbf{\Checkmark} & \multicolumn{2}{l}{1.510M} & \multicolumn{2}{l}{28.039} & \multicolumn{2}{l}{0.7718} \\ \hline
\textbf{\Checkmark} & \textbf{\XSolidBrush} & \multicolumn{2}{l}{1.537M} & \multicolumn{2}{l}{28.077} & \multicolumn{2}{l}{0.7723} \\ \hline
\textbf{\Checkmark} & \textbf{\Checkmark} & \multicolumn{2}{l}{1,659M} & \multicolumn{2}{l}{28.086} & \multicolumn{2}{l}{0.7732} \\ \hline
\end{tabular}%
}
\end{table}
\begin{figure}[tp]
    \centering 
    \includegraphics[width=1\linewidth]{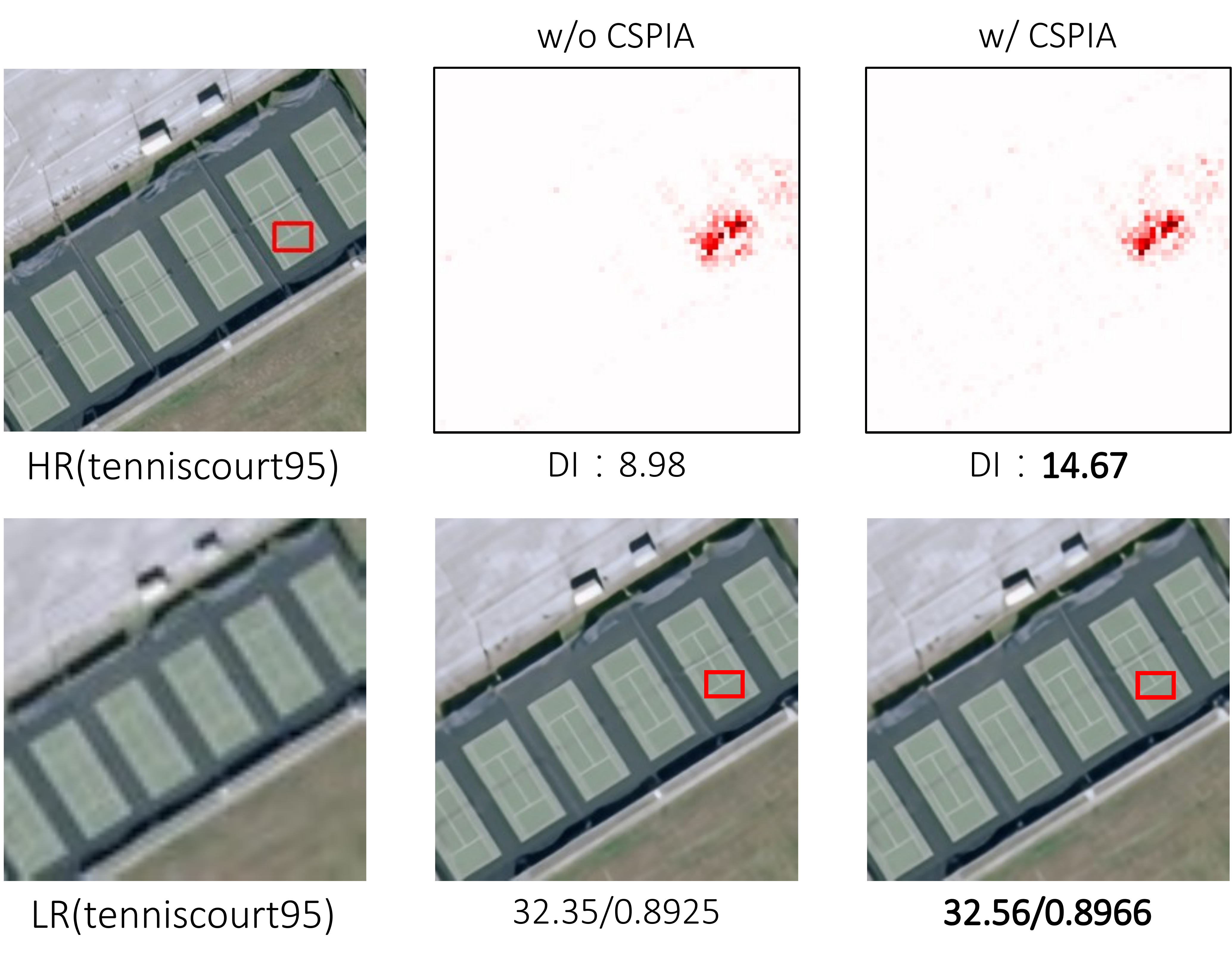}
    \caption{LAM \cite{lam} comparison between the full SPIFFNet (w/o CSPIA) and the variant without GPA (w/ CSPIA) for ×4 SR. The first column shows the low remote sensing resolution image and the corresponding high remote sensing resolution image. The second and third columns show SPIFFNet and its variants and the corresponding effective receptive fields.}
    \label{fig:ab}
\end{figure}
This article focuses on RSISR at three magnification factors: ×2, ×3, and ×4. During the training, we randomly sample LR remote sensing images and their corresponding HR reference windows as 48 × 48 windows. To augment the training samples, we apply random rotations (90°, 180°, and 270°) and horizontal flipping. The proposed SPIFFNet consists of 10 blocks, each of which is designed with a local window size and a global window size of 16 and a feature dimension of 64. Additionally, the SPIFFNet employs 4 attention heads. Further details and experimental analyses are presented in Section IV.

We employ the Adam optimizer \cite{kingma2014adam} for model optimization, setting $\beta_1$ = 0.9, $\beta_2$ = 0.99, and $\varepsilon$ = $10^{-8}$. We set the initial learning rate to $4 \times 10^{-4}$ and the mini-batch size to 8. We train the model for a total of 2000 epochs, gradually reducing the learning rate to $5 \times 10^{-7}$ at epoch 2000 using the cosine annealing schedule \cite{loshchilov2016sgdr}.
\section{EXPERIMENTAL RESULTS AND ANALYSES}
\begin{table*}[]
\centering
\caption{MEAN PSNR(DB) OF EACH CLASS FOR UPSCALING FACTOR 3
ON UCMERCED TEST DATASET}
\label{tab:uc}
\setlength{\tabcolsep}{4mm}{%
\begin{tabular}{ccclclclclclclcl}
\hline
\multirow{2}{*}{\textbf{Class NO.}} & \multirow{2}{*}{\textbf{Class name}} & \multicolumn{2}{c}{\multirow{2}{*}{\textbf{LGCNet}}} & \multicolumn{2}{c}{\multirow{2}{*}{\textbf{VDSR}}} & \multicolumn{2}{c}{\multirow{2}{*}{\textbf{DCM}}} & \multicolumn{2}{c}{\multirow{2}{*}{\textbf{HSENet}}} & \multicolumn{2}{c}{\multirow{2}{*}{\textbf{TransENet}}} & \multicolumn{2}{c}{\multirow{2}{*}{\textbf{\begin{tabular}[c]{@{}c@{}}SPIFFNet\\ (Ours)\end{tabular}}}} \\
 &  & \multicolumn{2}{c}{} & \multicolumn{2}{c}{} & \multicolumn{2}{c}{} & \multicolumn{2}{c}{} & \multicolumn{2}{c}{} & \multicolumn{2}{c}{} & \multicolumn{2}{c}{} \\ \hline
\textbf{1} & \textbf{agricultural} & \multicolumn{2}{c}{27.66} & \multicolumn{2}{c}{27.75} & \multicolumn{2}{c}{\textbf{29.06}} & \multicolumn{2}{c}{27.64} & \multicolumn{2}{c}{28.02} & \multicolumn{2}{c}{28.29} \\
\textbf{2} & \textbf{airplane} & \multicolumn{2}{c}{29.12} & \multicolumn{2}{c}{29.76} & \multicolumn{2}{c}{\textbf{30.77}} & \multicolumn{2}{c}{30.09} & \multicolumn{2}{c}{29.94} & \multicolumn{2}{c}{30.45} \\
\textbf{3} & \textbf{baseballdiamond} & \multicolumn{2}{c}{34.72} & \multicolumn{2}{c}{34.98} & \multicolumn{2}{c}{33.76} & \multicolumn{2}{c}{35.05} & \multicolumn{2}{c}{35.04} & \multicolumn{2}{c}{\textbf{35.61}} \\
\textbf{4} & \textbf{beach} & \multicolumn{2}{c}{37.37} & \multicolumn{2}{c}{37.57} & \multicolumn{2}{c}{36.38} & \multicolumn{2}{c}{37.69} & \multicolumn{2}{c}{37.53} & \multicolumn{2}{c}{\textbf{37.90}} \\
\textbf{5} & \textbf{buildings} & \multicolumn{2}{c}{27.81} & \multicolumn{2}{c}{28.53} & \multicolumn{2}{c}{28.51} & \multicolumn{2}{c}{28.95} & \multicolumn{2}{c}{28.81} & \multicolumn{2}{c}{\textbf{29.57}} \\
\textbf{6} & \textbf{chaparral} & \multicolumn{2}{c}{26.39} & \multicolumn{2}{c}{26.61} & \multicolumn{2}{c}{26.81} & \multicolumn{2}{c}{26.70} & \multicolumn{2}{c}{26.69} & \multicolumn{2}{c}{\textbf{26.87}} \\
\textbf{7} & \textbf{denseresidential} & \multicolumn{2}{c}{28.25} & \multicolumn{2}{c}{28.88} & \multicolumn{2}{c}{28.79} & \multicolumn{2}{c}{29.24} & \multicolumn{2}{c}{29.11} & \multicolumn{2}{c}{\textbf{29.66}} \\
\textbf{8} & \textbf{forest} & \multicolumn{2}{c}{28.44} & \multicolumn{2}{c}{28.52} & \multicolumn{2}{c}{28.16} & \multicolumn{2}{c}{28.59} & \multicolumn{2}{c}{28.59} & \multicolumn{2}{c}{\textbf{28.67}} \\
\textbf{9} & \textbf{freeway} & \multicolumn{2}{c}{29.52} & \multicolumn{2}{c}{30.21} & \multicolumn{2}{c}{30.45} & \multicolumn{2}{c}{30.63} & \multicolumn{2}{c}{30.38} & \multicolumn{2}{c}{\textbf{31.13}} \\
\textbf{10} & \textbf{golfcourse} & \multicolumn{2}{c}{36.51} & \multicolumn{2}{c}{36.57} & \multicolumn{2}{c}{34.43} & \multicolumn{2}{c}{36.62} & \multicolumn{2}{c}{36.68} & \multicolumn{2}{c}{\textbf{37.05}} \\
\textbf{11} & \textbf{harbor} & \multicolumn{2}{c}{23.63} & \multicolumn{2}{c}{24.36} & \multicolumn{2}{c}{\textbf{26.55}} & \multicolumn{2}{c}{24.88} & \multicolumn{2}{c}{24.72} & \multicolumn{2}{c}{25.47} \\
\textbf{12} & \textbf{intersection} & \multicolumn{2}{c}{28.29} & \multicolumn{2}{c}{28.83} & \multicolumn{2}{c}{29.28} & \multicolumn{2}{c}{29.21} & \multicolumn{2}{c}{29.03} & \multicolumn{2}{c}{\textbf{29.67}} \\
\textbf{13} & \textbf{mediumresidential} & \multicolumn{2}{c}{27.76} & \multicolumn{2}{c}{28.26} & \multicolumn{2}{c}{27.21} & \multicolumn{2}{c}{28.55} & \multicolumn{2}{c}{28.47} & \multicolumn{2}{c}{\textbf{28.94}} \\
\textbf{14} & \textbf{mobilehomepark} & \multicolumn{2}{c}{24.59} & \multicolumn{2}{c}{25.24} & \multicolumn{2}{c}{26.05} & \multicolumn{2}{c}{25.70} & \multicolumn{2}{c}{25.64} & \multicolumn{2}{c}{\textbf{26.25}} \\
\textbf{15} & \textbf{overpass} & \multicolumn{2}{c}{26.58} & \multicolumn{2}{c}{27.70} & \multicolumn{2}{c}{27.77} & \multicolumn{2}{c}{28.22} & \multicolumn{2}{c}{27.83} & \multicolumn{2}{c}{\textbf{28.55}} \\
\textbf{16} & \textbf{parkinglot} & \multicolumn{2}{c}{23.69} & \multicolumn{2}{c}{24.12} & \multicolumn{2}{c}{24.95} & \multicolumn{2}{c}{24.66} & \multicolumn{2}{c}{24.45} & \multicolumn{2}{c}{\textbf{25.37}} \\
\textbf{17} & \textbf{river} & \multicolumn{2}{c}{29.12} & \multicolumn{2}{c}{29.23} & \multicolumn{2}{c}{28.89} & \multicolumn{2}{c}{29.22} & \multicolumn{2}{c}{29.25} & \multicolumn{2}{c}{\textbf{29.54}} \\
\textbf{18} & \textbf{runway} & \multicolumn{2}{c}{31.15} & \multicolumn{2}{c}{31.41} & \multicolumn{2}{c}{\textbf{32.53}} & \multicolumn{2}{c}{31.15} & \multicolumn{2}{c}{31.25} & \multicolumn{2}{c}{31.69} \\
\textbf{19} & \textbf{sparseresidential} & \multicolumn{2}{c}{30.53} & \multicolumn{2}{c}{31.47} & \multicolumn{2}{c}{29.81} & \multicolumn{2}{c}{31.64} & \multicolumn{2}{c}{31.57} & \multicolumn{2}{c}{\textbf{31.89}} \\
\textbf{20} & \textbf{storagetanks} & \multicolumn{2}{c}{32.17} & \multicolumn{2}{c}{32.72} & \multicolumn{2}{c}{29.02} & \multicolumn{2}{c}{32.95} & \multicolumn{2}{c}{32.71} & \multicolumn{2}{c}{\textbf{33.23}} \\
\textbf{21} & \textbf{tenniscourt} & \multicolumn{2}{c}{31.58} & \multicolumn{2}{c}{32.29} & \multicolumn{2}{c}{30.76} & \multicolumn{2}{c}{32.71} & \multicolumn{2}{c}{32.51} & \multicolumn{2}{c}{\textbf{33.19}} \\ \hline
\multicolumn{2}{c}{\textbf{avg}} & \multicolumn{2}{c}{29.28} & \multicolumn{2}{c}{29.76} & \multicolumn{2}{c}{29.52} & \multicolumn{2}{c}{30.00} & \multicolumn{2}{c}{29.92} & \multicolumn{2}{c}{\textbf{30.43}} \\ \hline
\end{tabular}%
}
\end{table*}
\subsection{Experimental Datasets and Metrics}
1) \emph{Datasets:} This study employs two publicly available remote sensing datasets, namely UCMecred \cite{uc} and AID \cite{aid}. The UCMerced dataset comprises 21 classes, each containing 100 remote sensing scenes images that have dimensions of 256 × 256 pixels. We partitioned the dataset into two subsets: one for training purposes and the other for testing. Each subset comprises 1050 images. The AID dataset consists of 10,000 images representing 30 classes of remote sensing scenes which have dimensions of 600 × 600 pixels. In the case of the AID dataset, 80$\%$ of the total dataset is randomly assigned as the training set, while the remaining images are allocated for testing.

2) \emph{Metrics:} We select PSNR and SSIM \cite{wang2004image} as the evaluation metrics for RSISR, and assess all super-resolution results on the RGB channels. 
\subsection{Ablation Studies}
We conducted experiments in this section to validate the components of our method. All experiments were performed using the same experimental setup, with the UCMereced dataset and a uniform magnification factor of 4 was applied. We start with a naive baseline by removing both components. Then we add CSPIA and CSFFA to the baseline, respectively. At last, both components are employed to compose our final version of method. The results are reported in Table \ref{tab:Ablation Studies}.

\emph{1) Effects of CSPIA:} Table \ref{tab:Ablation Studies} summarizes the results of this ablation study. We can see that the model with CSPIA significantly outperforms the baseline model, indicating the effectiveness of the expandable window mechanism in capturing both global and local features.

To better understand the main reason of the improvement brought by CSPIA, we utilize LAM \cite{lam} to visualize the effective receptive field of a input window. As shown in Fig. \ref{fig:ab}, the window benefits from a global range of useful pixels by using CSPIA. The results indicate the effectiveness of the proposed CSPIA in improving PSNR and SSIM performances.

\emph{2) Effects of CSFFA:} Table \ref{tab:Ablation Studies} summarizes the results of this ablation study. The model incorporating CSFFA demonstrates a significant performance improvement over the baseline model, providing strong evidence of the effectiveness of CSFFA.
\subsection{Comparisons with Other Methods}
\begin{table*}[]
\centering
\caption{MEAN PSNR(DB) OF EACH CLASS FOR UPSCALING FACTOR 4 ON AID
TEST DATASET}
\label{tab:aid}
\setlength{\tabcolsep}{4mm}{%
\begin{tabular}{ccclclclclclclcl}
\hline
\multirow{2}{*}{\textbf{Class NO.}} & \multirow{2}{*}{\textbf{Class Name}} & \multicolumn{2}{c}{\multirow{2}{*}{\textbf{LGCNet}}} & \multicolumn{2}{c}{\multirow{2}{*}{\textbf{VDSR}}} & \multicolumn{2}{c}{\multirow{2}{*}{\textbf{DCM}}} & \multicolumn{2}{c}{\multirow{2}{*}{\textbf{HSENet}}} & \multicolumn{2}{c}{\multirow{2}{*}{\textbf{TransENet}}} & \multicolumn{2}{c}{\multirow{2}{*}{\textbf{\begin{tabular}[c]{@{}c@{}}SPIFFNet\\ (Ours)\end{tabular}}}} \\
 &  & \multicolumn{2}{c}{} & \multicolumn{2}{c}{} & \multicolumn{2}{c}{} & \multicolumn{2}{c}{} & \multicolumn{2}{c}{} & \multicolumn{2}{c}{} & \multicolumn{2}{c}{} \\ \hline
\textbf{1} & \textbf{airport} & \multicolumn{2}{c}{28.39} & \multicolumn{2}{c}{28.82} & \multicolumn{2}{c}{28.99} & \multicolumn{2}{c}{29.15} & \multicolumn{2}{c}{29.23} &  \multicolumn{2}{c}{\textbf{29.34}} \\
\textbf{2} & \textbf{bareland} & \multicolumn{2}{c}{35.78} & \multicolumn{2}{c}{35.98} & \multicolumn{2}{c}{36.17} & \multicolumn{2}{c}{36.25} & \multicolumn{2}{c}{36.20} &  \multicolumn{2}{c}{\textbf{36.48}} \\
\textbf{3} & \textbf{baseballfield} & \multicolumn{2}{c}{30.75} & \multicolumn{2}{c}{31.18} & \multicolumn{2}{c}{31.36} & \multicolumn{2}{c}{31.52} & \multicolumn{2}{c}{\textbf{31.59}} &  \multicolumn{2}{c}{31.48} \\
\textbf{4} & \textbf{beach} & \multicolumn{2}{c}{32.08} & \multicolumn{2}{c}{32.29} & \multicolumn{2}{c}{32.45} & \multicolumn{2}{c}{32.54} & \multicolumn{2}{c}{32.55} &  \multicolumn{2}{c}{\textbf{32.78}} \\
\textbf{5} & \textbf{bridge} & \multicolumn{2}{c}{30.67} & \multicolumn{2}{c}{31.19} & \multicolumn{2}{c}{31.39} & \multicolumn{2}{c}{31.57} & \multicolumn{2}{c}{31.63} &  \multicolumn{2}{c}{\textbf{31.87}} \\
\textbf{6} & \textbf{center} & \multicolumn{2}{c}{26.92} & \multicolumn{2}{c}{27.48} & \multicolumn{2}{c}{27.72} & \multicolumn{2}{c}{27.95} & \multicolumn{2}{c}{28.03} &  \multicolumn{2}{c}{\textbf{28.21}} \\
\textbf{7} & \textbf{church} & \multicolumn{2}{c}{23.68} & \multicolumn{2}{c}{24.12} & \multicolumn{2}{c}{24.29} & \multicolumn{2}{c}{24.47} & \multicolumn{2}{c}{24.51} &  \multicolumn{2}{c}{\textbf{24.63}} \\
\textbf{8} & \textbf{commercial} & \multicolumn{2}{c}{27.24} & \multicolumn{2}{c}{27.62} & \multicolumn{2}{c}{27.78} & \multicolumn{2}{c}{27.94} & \multicolumn{2}{c}{27.97} &  \multicolumn{2}{c}{\textbf{28.07}} \\
\textbf{9} & \textbf{denseresidential} & \multicolumn{2}{c}{24.33} & \multicolumn{2}{c}{24.70} & \multicolumn{2}{c}{24.87} & \multicolumn{2}{c}{25.06} & \multicolumn{2}{c}{25.13} &  \multicolumn{2}{c}{\textbf{25.20}} \\
\textbf{10} & \textbf{desert} & \multicolumn{2}{c}{39.06} & \multicolumn{2}{c}{39.13} & \multicolumn{2}{c}{39.27} & \multicolumn{2}{c}{39.37} & \multicolumn{2}{c}{39.31} &  \multicolumn{2}{c}{\textbf{39.58}} \\
\textbf{11} & \textbf{farmland} & \multicolumn{2}{c}{33.77} & \multicolumn{2}{c}{34.20} & \multicolumn{2}{c}{34.42} & \multicolumn{2}{c}{34.56} & \multicolumn{2}{c}{34.58} &  \multicolumn{2}{c}{\textbf{34.79}} \\
\textbf{12} & \textbf{forest} & \multicolumn{2}{c}{28.20} & \multicolumn{2}{c}{28.36} & \multicolumn{2}{c}{28.47} & \multicolumn{2}{c}{28.56} & \multicolumn{2}{c}{28.56} &  \multicolumn{2}{c}{\textbf{28.62}} \\
\textbf{13} & \textbf{industrial} & \multicolumn{2}{c}{26.24} & \multicolumn{2}{c}{26.72} & \multicolumn{2}{c}{26.92} & \multicolumn{2}{c}{27.13} & \multicolumn{2}{c}{27.21} &  \multicolumn{2}{c}{\textbf{27.32}} \\
\textbf{14} & \textbf{meadow} & \multicolumn{2}{c}{32.65} & \multicolumn{2}{c}{32.77} & \multicolumn{2}{c}{32.88} & \multicolumn{2}{c}{32.94} & \multicolumn{2}{c}{32.94} &  \multicolumn{2}{c}{\textbf{33.14}} \\
\textbf{15} & \textbf{mediumresidential} & \multicolumn{2}{c}{27.63} & \multicolumn{2}{c}{28.06} & \multicolumn{2}{c}{28.25} & \multicolumn{2}{c}{28.44} & \multicolumn{2}{c}{28.45} &  \multicolumn{2}{c}{\textbf{28.57}} \\
\textbf{16} & \textbf{mountain} & \multicolumn{2}{c}{28.97} & \multicolumn{2}{c}{29.11} & \multicolumn{2}{c}{29.18} & \multicolumn{2}{c}{29.22} & \multicolumn{2}{c}{29.28} &  \multicolumn{2}{c}{\textbf{29.33}} \\
\textbf{17} & \textbf{park} & \multicolumn{2}{c}{27.37} & \multicolumn{2}{c}{27.69} & \multicolumn{2}{c}{27.82} & \multicolumn{2}{c}{27.95} & \multicolumn{2}{c}{28.01} &  \multicolumn{2}{c}{\textbf{28.16}} \\
\textbf{18} & \textbf{parking} & \multicolumn{2}{c}{24.40} & \multicolumn{2}{c}{25.21} & \multicolumn{2}{c}{25.74} & \multicolumn{2}{c}{26.27} & \multicolumn{2}{c}{26.40} &  \multicolumn{2}{c}{\textbf{26.56}} \\
\textbf{19} & \textbf{playground} & \multicolumn{2}{c}{29.04} & \multicolumn{2}{c}{29.62} & \multicolumn{2}{c}{29.92} & \multicolumn{2}{c}{30.20} & \multicolumn{2}{c}{30.30} &  \multicolumn{2}{c}{\textbf{30.64}} \\
\textbf{20} & \textbf{pond} & \multicolumn{2}{c}{30.00} & \multicolumn{2}{c}{30.26} & \multicolumn{2}{c}{30.39} & \multicolumn{2}{c}{30.49} & \multicolumn{2}{c}{30.53} &  \multicolumn{2}{c}{\textbf{30.86}} \\
\textbf{21} & \textbf{port} & \multicolumn{2}{c}{26.02} & \multicolumn{2}{c}{26.43} & \multicolumn{2}{c}{26.62} & \multicolumn{2}{c}{26.84} & \multicolumn{2}{c}{26.91} &  \multicolumn{2}{c}{\textbf{26.99}} \\
\textbf{22} & \textbf{railwaystation} & \multicolumn{2}{c}{27.76} & \multicolumn{2}{c}{28.19} & \multicolumn{2}{c}{28.38} & \multicolumn{2}{c}{28.55} & \multicolumn{2}{c}{28.61} &  \multicolumn{2}{c}{\textbf{28.74}} \\
\textbf{23} & \textbf{resort} & \multicolumn{2}{c}{27.32} & \multicolumn{2}{c}{27.71} & \multicolumn{2}{c}{27.88} & \multicolumn{2}{c}{28.04} & \multicolumn{2}{c}{28.08} &  \multicolumn{2}{c}{\textbf{28.13}} \\
\textbf{24} & \textbf{river} & \multicolumn{2}{c}{30.60} & \multicolumn{2}{c}{30.82} & \multicolumn{2}{c}{30.91} & \multicolumn{2}{c}{30.98} & \multicolumn{2}{c}{31.00} &  \multicolumn{2}{c}{\textbf{31.11}} \\
\textbf{25} & \textbf{school} & \multicolumn{2}{c}{26.34} & \multicolumn{2}{c}{26.78} & \multicolumn{2}{c}{26.94} & \multicolumn{2}{c}{27.15} & \multicolumn{2}{c}{27.22} &  \multicolumn{2}{c}{\textbf{27.35}} \\
\textbf{26} & \textbf{sparseresidential} & \multicolumn{2}{c}{26.27} & \multicolumn{2}{c}{26.46} & \multicolumn{2}{c}{26.53} & \multicolumn{2}{c}{26.63} & \multicolumn{2}{c}{26.63} &  \multicolumn{2}{c}{\textbf{26.68}} \\
\textbf{27} & \textbf{square} & \multicolumn{2}{c}{28.39} & \multicolumn{2}{c}{28.91} & \multicolumn{2}{c}{29.13} & \multicolumn{2}{c}{29.33} & \multicolumn{2}{c}{29.39} &  \multicolumn{2}{c}{\textbf{29.59}} \\
\textbf{28} & \textbf{stadium} & \multicolumn{2}{c}{26.37} & \multicolumn{2}{c}{26.88} & \multicolumn{2}{c}{27.10} & \multicolumn{2}{c}{27.32} & \multicolumn{2}{c}{27.41} &  \multicolumn{2}{c}{\textbf{27.54}} \\
\textbf{29} & \textbf{storagetanks} & \multicolumn{2}{c}{25.48} & \multicolumn{2}{c}{25.86} & \multicolumn{2}{c}{26.00} & \multicolumn{2}{c}{26.16} & \multicolumn{2}{c}{26.20} &  \multicolumn{2}{c}{\textbf{26.26}} \\
\textbf{30} & \textbf{viaduct} & \multicolumn{2}{c}{27.26} & \multicolumn{2}{c}{27.74} & \multicolumn{2}{c}{27.93} & \multicolumn{2}{c}{28.13} & \multicolumn{2}{c}{28.21} &  \multicolumn{2}{c}{\textbf{28.36}} \\ \hline
\multicolumn{2}{c}{\textbf{avg}} & \multicolumn{2}{c}{28.61} & \multicolumn{2}{c}{28.99} & \multicolumn{2}{c}{29.17} & \multicolumn{2}{c}{29.34} & \multicolumn{2}{c}{29.38} &  \multicolumn{2}{c}{\textbf{29.54}} \\ \hline
\end{tabular}%
}
\end{table*}
This section presents a comparative analysis of the proposed method with several deep learning-based SR methods, namely SRCNN \cite{srcnn}, VDSR \cite{VDSR}, LGCNet \cite{LGCnet}, DCM \cite{dcm}, HSENet \cite{hsenet}, and TransENet \cite{transenet}. 
\begin{figure*}[htp]
    \centering 
    \includegraphics[width=1\linewidth]{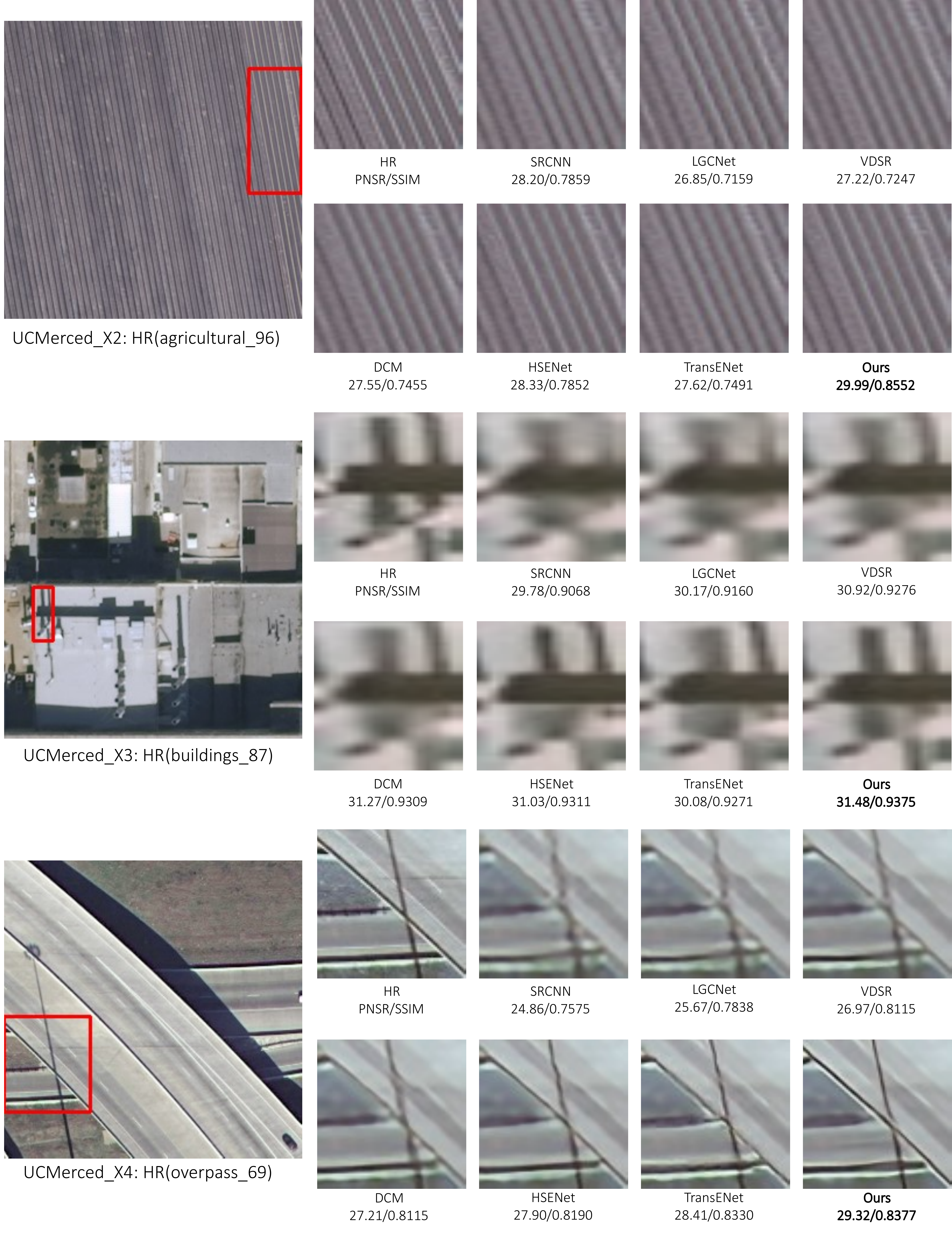}
    \caption{Result comparisons on UCMerced dataset with different methods.}
    \label{fig:uc}
\end{figure*}
\begin{figure*}[htp]
    \centering 
    \includegraphics[width=1\linewidth]{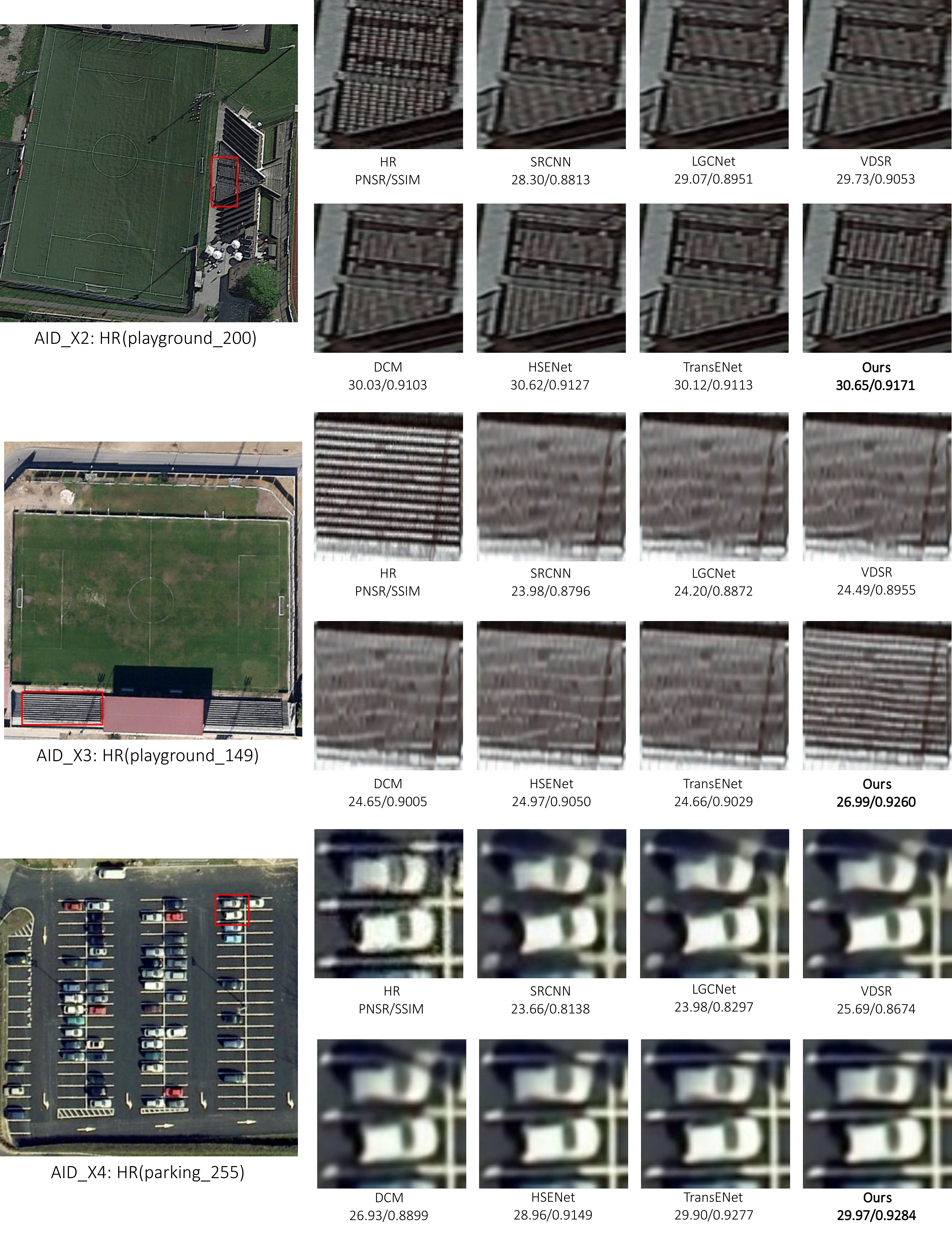}
    \caption{Result comparisons on AID dataset with different methods.}
    \label{fig:aid}
\end{figure*}

\emph{1) Quantitative Results on UCMerced Dataset:} The performance of these approachs on the UCMerced dataset is reported in Table \ref{tab:all}. The best result is indicated in bold font. Notably, some results have been reported in multiple published articles \cite{dcm}, \cite{qin2020remote}. To ensure consistency, we retrained these comparison methods using the open-source code, subjecting all methods to the same testing conditions. The results demonstrate that our SPIFFNet achieves the highest values in terms of PSNR and SSIM. Table \ref{tab:uc} provides a summary of the PSNR for each class in the UCMeced dataset at an upscale factor of 3. We observed that SPIFFNet significantly outperforms HSENet \cite{hsenet} and TransENet \cite{transenet} on the buildings, harbors and parking lot classes, which require precise local context information for object discrimination and image detail reconstruction. These notable results serve as further evidence of the effectiveness of our proposed method.

\emph{2) Quantitative Results on AID Dataset:} We conducted additional experiments on the AID dataset to further validate the effectiveness of SPIFFNet. Table \ref{tab:all} presents the PSNR and SSIM results of SPIFFNet compared to other methods on this dataset. Compared to other methods, SPIFFNet achieves the best results. Furthermore, Table \cite{aid} presents the test results for each category at a magnification factor of 4 on the AID dataset. The consistent superior performance of SPIFF in various scenarios further demonstrates the effectiveness of our approach.

\emph{3) Qualitative Results:} Fig. \ref{fig:uc} displays several super-resolved examples from the UCMerced dataset, such as scenes depicting "agricultural", "buildings", and "overpass". Similarly, Fig. \ref{fig:aid} showcases examples from the AID dataset, including scenes of "playground" and "parking". Our method demonstrates superior performance compared to other methods in challenging areas such as texture and edge, as evident from the visual results. This observation provides further evidence of the effectiveness of our approach.
\section{CONCLUSION}
This paper introduces a novel transformer-based method called Cross-Spatial Pixel Integration and Cross-Stage Feature Fusion Based Transformer Network (SPIFFNet) for RSISR. The aim of SPIFFNet is to enhance the global perception ability of local windows by introducing context information and to improve the representation ability of features by integrating cross-stage features. SPIFFNet consists of two key components: CSPIA, which introduces context information into the reconstruction of local windows to enhance global awareness, and CSFFA, which enables adaptive aggregation of features across different stages of the network, resulting in more effective information fusion and superior super-resolution performance. We conducted extensive experiments on benchmark datasets to validate the effectiveness of SPIFFNet. 

\bibliographystyle{IEEEtran}
\bibliography{remoteref}

\begin{IEEEbiography}[{\includegraphics[width=1in,height=1.25in,clip,keepaspectratio]{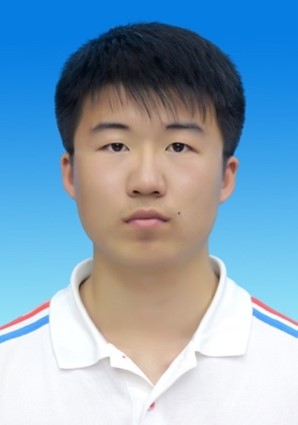}}]{Yuting Lu} is pursuing the Ph.D. degree in control science and engineering from the Northwestern Polytechnic University,Xi’an, China. His research interests include image super-resolution and computer vision.
\end{IEEEbiography}

\begin{IEEEbiography}[{\includegraphics[width=1in,height=1.25in,clip,keepaspectratio]{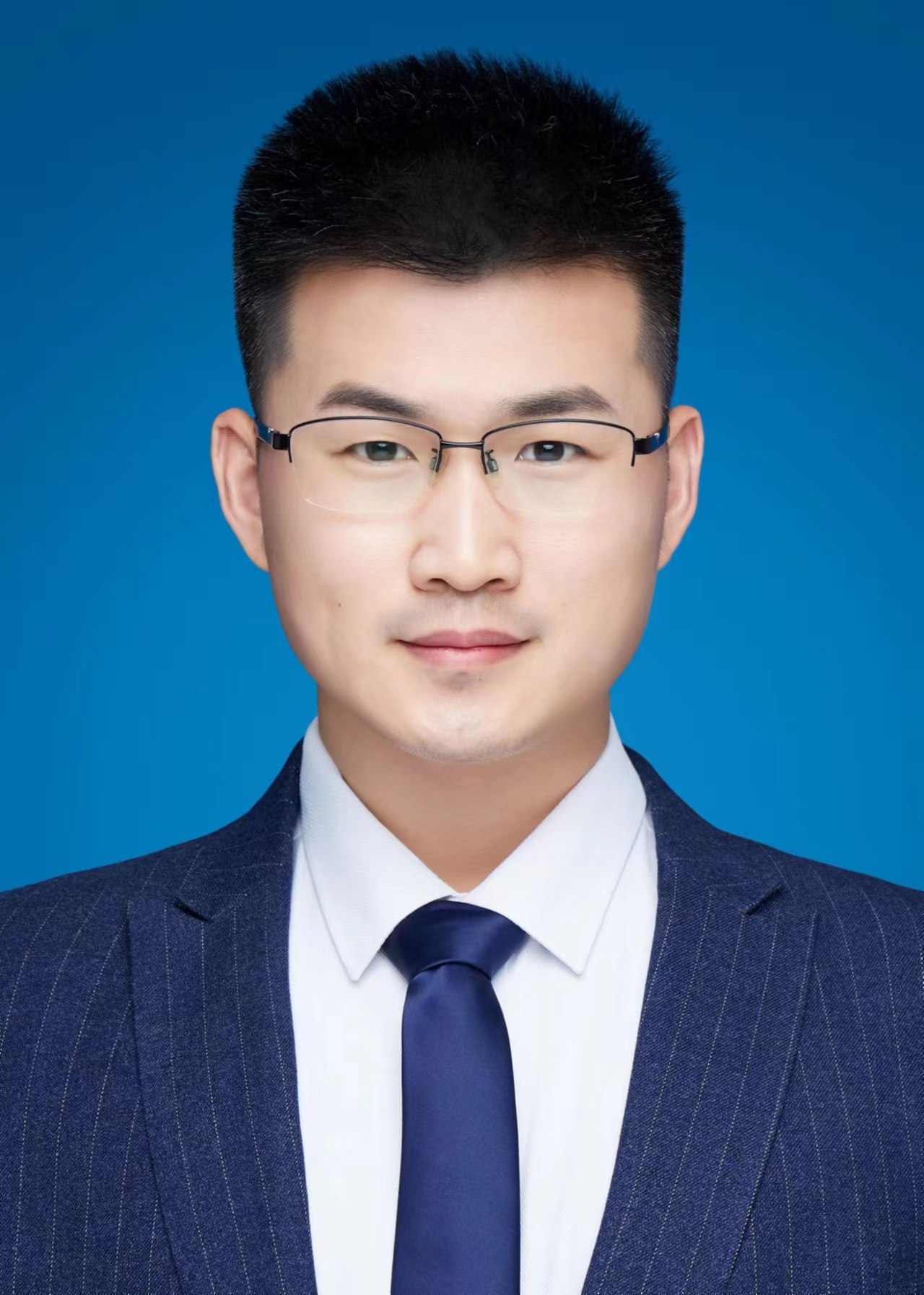}}]{Lingtong Min} received the B.S. degree from Northeastern University, Shenyang, China, in 2012, and the Ph.D. degree from Zhejiang University, Hangzhou, in 2019. He is an Associate Professor with Northwestern Polytechnical University. His main research interests are computer vision, pattern recognition, and remote sensing image understanding.
\end{IEEEbiography}

\begin{IEEEbiography}[{\includegraphics[width=1in,height=1.25in,clip,keepaspectratio]{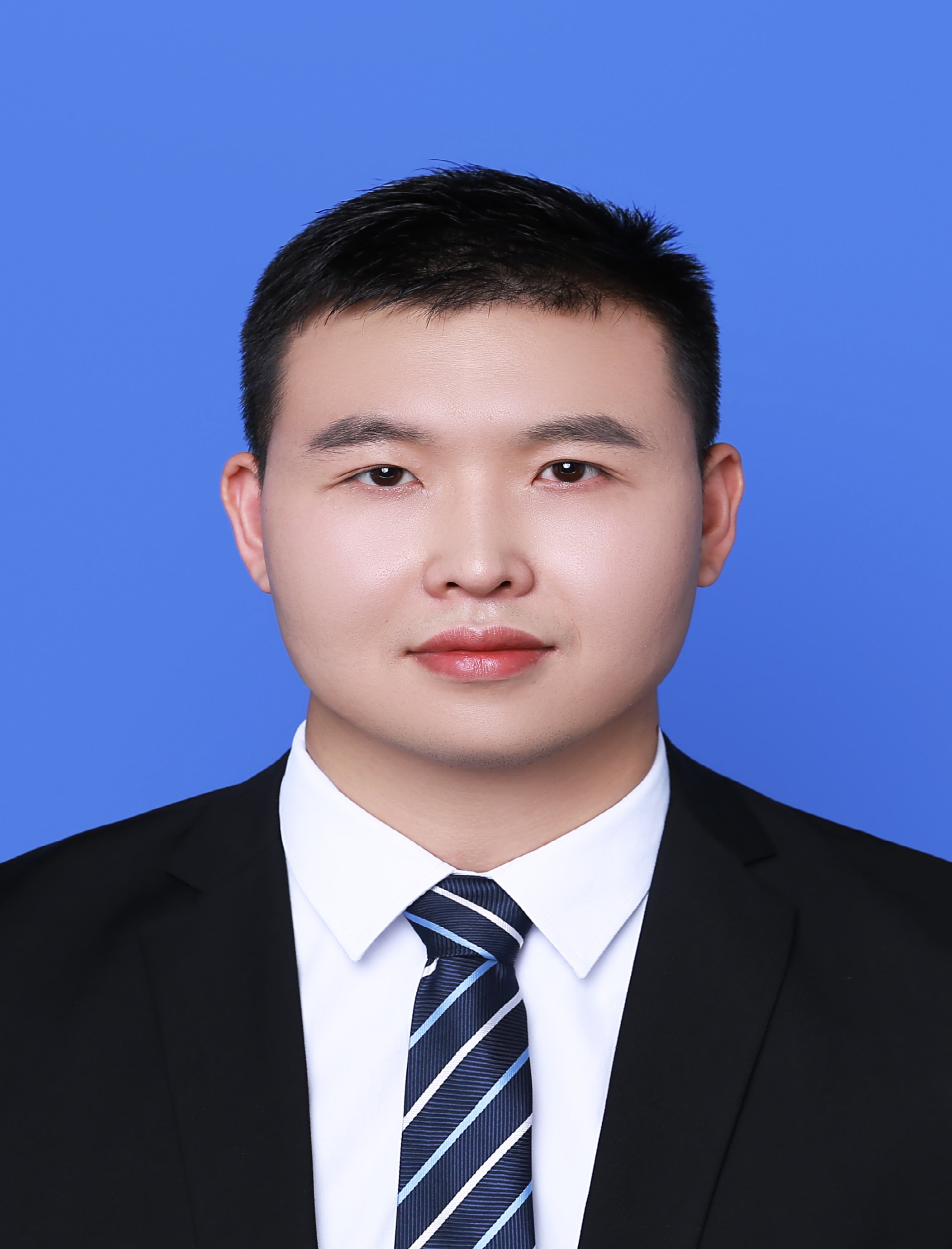}}]{Binglu Wang} (M’21) received the Ph.D. degree in Control Science and Engineering with the School of Automation at Northwestern Polytechnic University, Xi'an, China, in 2021. He is currently a Post-doctoral with the Department of Electrical Engineering, Beijing Institute of Technology, Beijing, China. His research interests include Computer Vision, Digital Signal Processing and Deep Learning.
\end{IEEEbiography}

\begin{IEEEbiography}[{\includegraphics[width=1in,height=1.25in,clip,keepaspectratio]{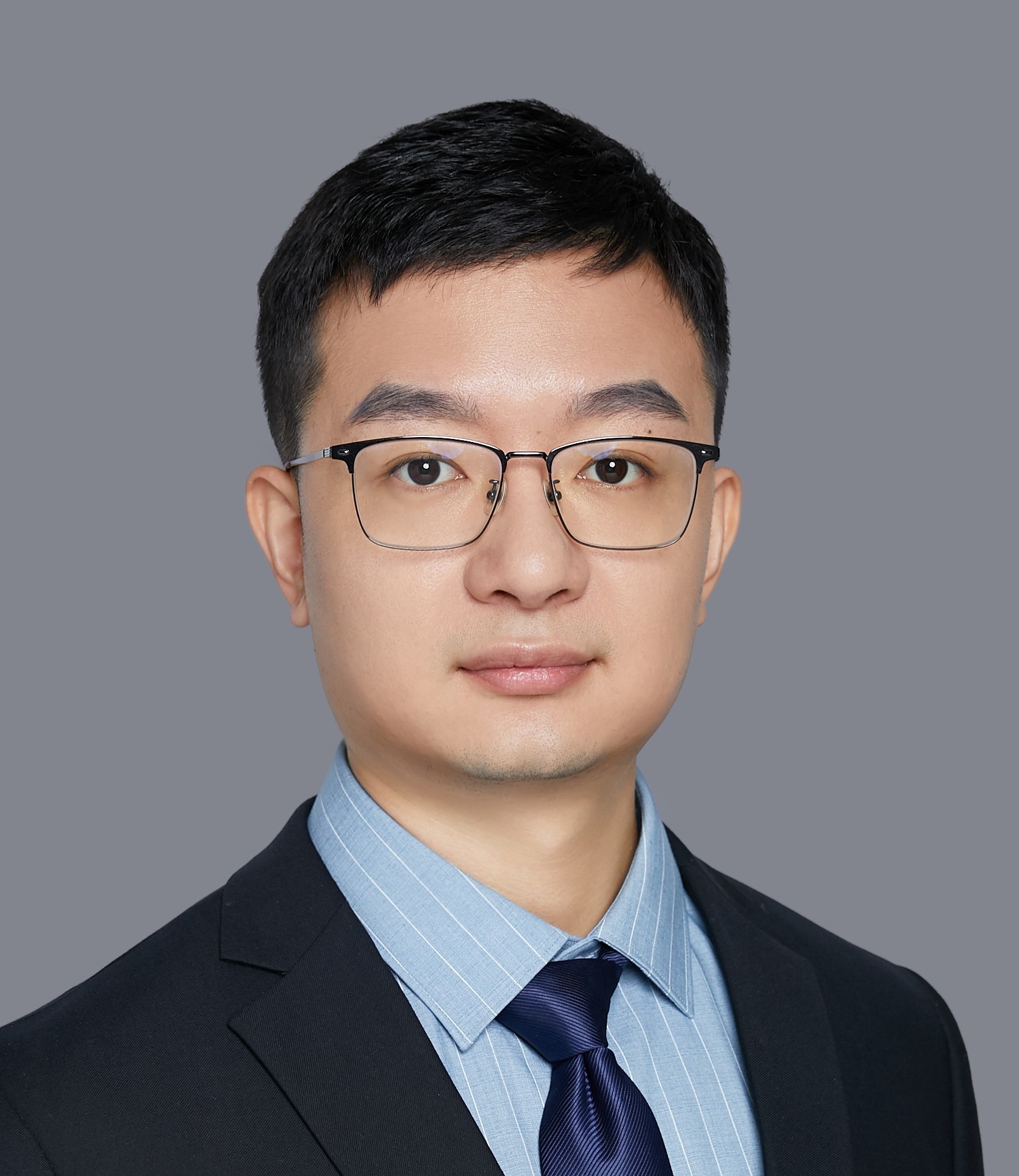}}]{Le Zheng} (Senior Member, IEEE) received the B.Eng. degree from Northwestern Polytechnical University (NWPU), Xi’an, China, in 2009 and Ph.D degree from Beijing Institute of Technology (BIT), Beijing, China in 2015, respectively. He has previously held academic positions in the Electrical Engineering Department of Columbia University, New York, U.S., first as a Visiting Researcher from 2013 to 2014 and then as a Postdoc Research Fellow from 2015 to 2017. From 2018 to 2022, he worked at Aptiv (formerly Delphi), Los Angeles, as a Principal Radar Systems Engineer, leading projects on the next-generation automotive radar products. Since July 2022, he has been a Full Professor with the School of Information and Electronics, BIT. His research interests lie in the general areas of radar, statistical signal processing, wireless communication, and high-performance hardware, and in particular in the area of automotive radar and integrated sensing and communications (ISAC).
\end{IEEEbiography}

\begin{IEEEbiography}[{\includegraphics[width=1in,height=1.25in,clip,keepaspectratio]{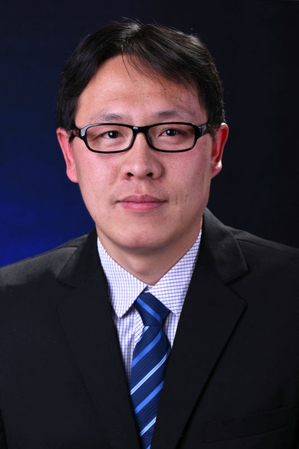}}]{Xiaoxu Wang}(M’10) received the M.S. and Ph.D. degrees from the School of Automation, Harbin Engineering University, Harbin, China, in 2008 and 2010, respectively. He was as a Postdoctoral Researcher from 2010 to 2012, and as an Associate Professor from 2013 to October 2018 in Automation School of Northwestern Polytechnical University. He is currently a professor with the Northwestern Polytechnical University. His main research interests include deep learning, inertial navigation and nonlinear estimation.
\end{IEEEbiography}

\begin{IEEEbiography}[{\includegraphics[width=1in,height=1.25in,clip,keepaspectratio]{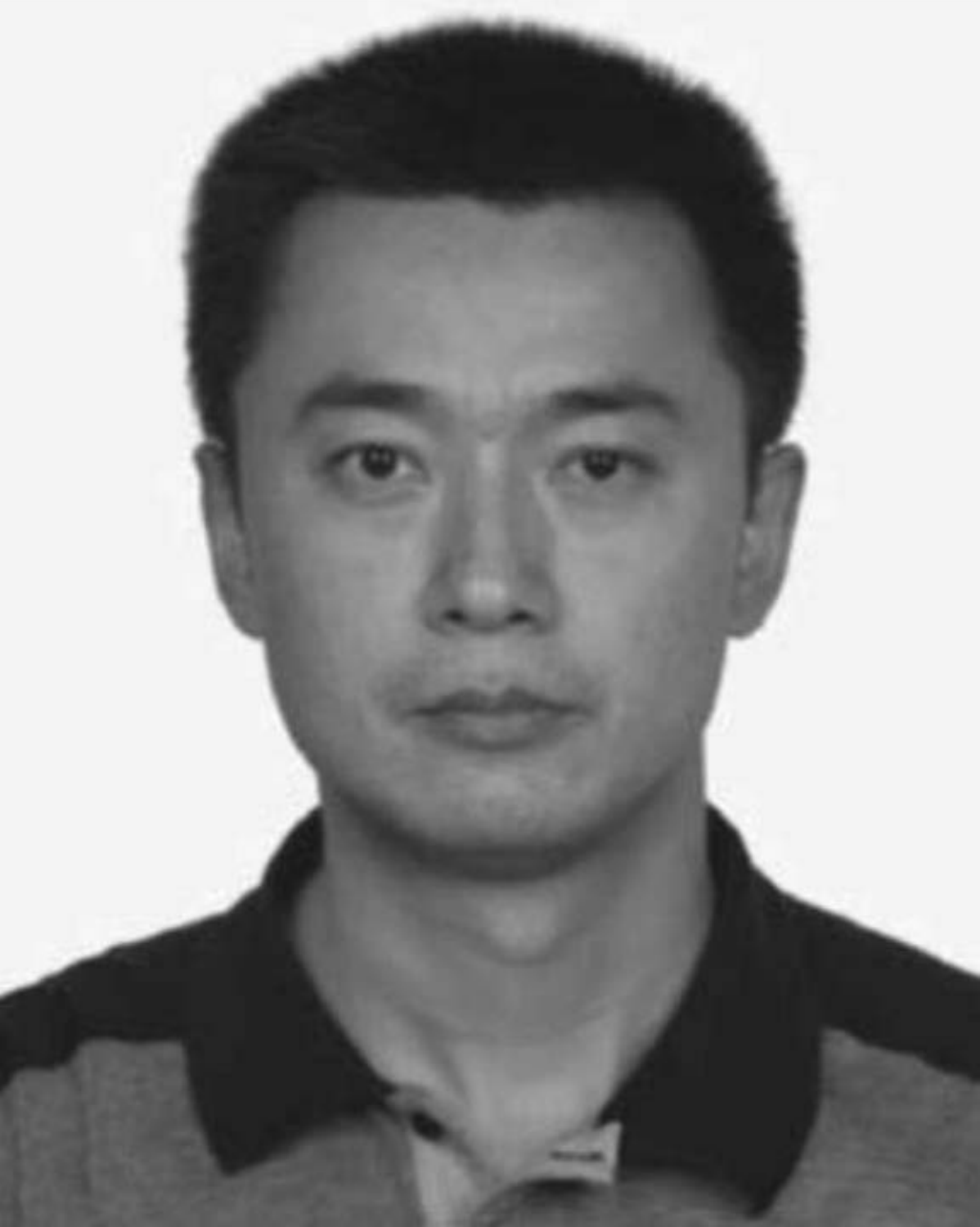}}]{Yongqiang Zhao}(M’05) received the B.S., M.S., and Ph.D. degrees in control science and engineering from the Northwestern Polytechnic University,Xi’an, China.From 2007 to 2009, he was as a Post-Doctora Researcher with McMaster University, Hamilton, ON, Canada, and Temple University, Philadelphia,PA, USA. He is currently a Professor with the Northwestern Polytechnical University. His research interests include polarization vision, hyperspectral imaging, and pattern recognition.
\end{IEEEbiography}	

\begin{IEEEbiography}[{\includegraphics[width=1in,height=1.25in,clip,keepaspectratio]{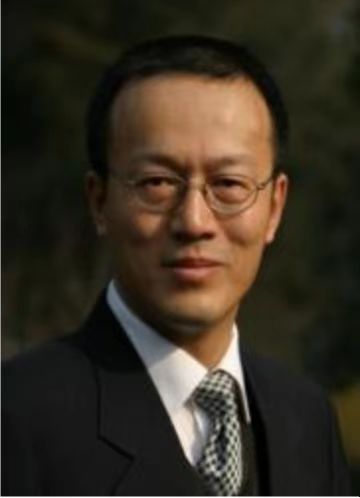}}]{Teng Long} (Fellow IEEE) was born in Fujian, China, in 1968. He received the M.S. and. Ph.D. degrees in electrical engineering from the Beijing Institute of Technology, Beijing, China, in 1991 and 1995, respectively. He was a Visiting Scholar with Stanford University, California, in 1999, and University College London, in 2002. He has been a Full Professor with the Department of Electrical Engineering, Beijing Institute of Technology, since 2000. He has authored or co-authored more than 300 articles. His research interests include synthetic aperture radar systems and real-time digital signal processing, with applications to radar and communication systems. Dr. Long is a Fellow of the Institute of Electronic and Technology and the Chinese Institute of Electronics. He was the recipient of many awards for his contributions to research and invention in China. He has been a member of the Chinese Engineering Academy since 2021.
\end{IEEEbiography}

\end{document}